\documentclass[10pt,twocolumn,letterpaper]{article}

\usepackage[pagenumbers]{cvpr2023_stylekit/cvpr} 

\usepackage{graphicx}
\usepackage{amsmath}
\usepackage{amssymb}
\usepackage{booktabs}
\usepackage{bm}
\usepackage[dvipsnames,table,xcdraw]{xcolor}
\usepackage{rotating}
\usepackage{algorithm}
\usepackage{algpseudocode}

\usepackage[pagebackref,breaklinks,colorlinks]{hyperref}

\usepackage[capitalize]{cleveref}
\crefname{section}{Sec.}{Secs.}
\Crefname{section}{Section}{Sections}
\Crefname{table}{Table}{Tables}
\crefname{table}{Tab.}{Tabs.} %

\def\etal{\emph{et al}\onedot}

\begin{document}

\title{Clockwork Diffusion: Efficient Generation With Model-Step Distillation}

\author{Amirhossein Habibian\thanks{Equal contribution} \qquad
Amir Ghodrati\footnotemark[1] \qquad
Noor Fathima\footnotemark[1] \qquad
Guillaume Sautiere \\ 
Risheek Garrepalli \qquad 
Fatih Porikli \qquad
Jens Petersen \\
{Qualcomm AI Research\thanks{Qualcomm AI Research is an initiative of Qualcomm Technologies, Inc}} \\
{\tt\small \{ahabibia, ghodrati, noor, gsautie, rgarrepa, fporikli, jpeterse\}@qti.qualcomm.com}
}



\newcommand{\head}[1]{{\smallskip\noindent\textbf{#1}}}
\newcommand{\alert}[1]{{\color{red}{#1}}}
\newcommand{\sm}{\scriptsize}
\newcommand{\eq}[1]{(\ref{eq:#1})}

\newcommand{\Th}[1]{\textsc{#1}}
\newcommand{\mr}[2]{\multirow{#1}{*}{#2}}
\newcommand{\mc}[2]{\multicolumn{#1}{c}{#2}}
\newcommand{\tb}[1]{\textbf{#1}}
\newcommand{\ul}[1]{\underline{#1}}
\newcommand{\ch}{\checkmark}

\newcommand{\red}[1]{{\color{red}{#1}}}
\newcommand{\blue}[1]{{\color{blue}{#1}}}
\newcommand{\green}[1]{\color{green}{#1}}
\newcommand{\gray}[1]{{\color{gray}{#1}}}

\newcommand{\citeme}[1]{\red{[XX]}}
\newcommand{\refme}[1]{\red{(XX)}}

\newcommand{\fig}[2][1]{\includegraphics[width=#1\linewidth]{fig/#2}}
\newcommand{\figh}[2][1]{\includegraphics[height=#1\linewidth]{fig/#2}}


\newcommand{\tran}{^\top}
\newcommand{\mtran}{^{-\top}}
\newcommand{\zcol}{\mathbf{0}}
\newcommand{\zrow}{\zcol\tran}

\newcommand{\ind}{\mathbbm{1}}
\newcommand{\expect}{\mathbb{E}}
\newcommand{\nat}{\mathbb{N}}
\newcommand{\zahl}{\mathbb{Z}}
\newcommand{\real}{\mathbb{R}}
\newcommand{\proj}{\mathbb{P}}
\newcommand{\prob}{\mathbf{Pr}}
\newcommand{\normal}{\mathcal{N}}

\newcommand{\mif}{\textrm{if}\ }
\newcommand{\other}{\textrm{otherwise}}
\newcommand{\minimize}{\textrm{minimize}\ }
\newcommand{\maximize}{\textrm{maximize}\ }
\newcommand{\st}{\textrm{subject\ to}\ }

\newcommand{\id}{\operatorname{id}}
\newcommand{\const}{\operatorname{const}}
\newcommand{\sgn}{\operatorname{sgn}}
\newcommand{\var}{\operatorname{Var}}
\newcommand{\mean}{\operatorname{mean}}
\newcommand{\trace}{\operatorname{tr}}
\newcommand{\diag}{\operatorname{diag}}
\newcommand{\vect}{\operatorname{vec}}
\newcommand{\cov}{\operatorname{cov}}
\newcommand{\sign}{\operatorname{sign}}
\newcommand{\prj}{\operatorname{proj}}

\newcommand{\softmax}{\operatorname{softmax}}
\newcommand{\clip}{\operatorname{clip}}

\newcommand{\defn}{\mathrel{:=}}
\newcommand{\peq}{\mathrel{+\!=}}
\newcommand{\meq}{\mathrel{-\!=}}

\newcommand{\floor}[1]{\left\lfloor{#1}\right\rfloor}
\newcommand{\ceil}[1]{\left\lceil{#1}\right\rceil}
\newcommand{\inner}[1]{\left\langle{#1}\right\rangle}
\newcommand{\norm}[1]{\left\|{#1}\right\|}
\newcommand{\abs}[1]{\left|{#1}\right|}
\newcommand{\frob}[1]{\norm{#1}_F}
\newcommand{\card}[1]{\left|{#1}\right|\xspace}
\newcommand{\divg}[2]{{#1\ ||\ #2}}
\newcommand{\diff}{\mathrm{d}}
\newcommand{\der}[3][]{\frac{d^{#1}#2}{d#3^{#1}}}
\newcommand{\pder}[3][]{\frac{\partial^{#1}{#2}}{\partial{#3^{#1}}}}
\newcommand{\ipder}[3][]{\partial^{#1}{#2}/\partial{#3^{#1}}}
\newcommand{\dder}[3]{\frac{\partial^2{#1}}{\partial{#2}\partial{#3}}}

\newcommand{\wb}[1]{\overline{#1}}
\newcommand{\wt}[1]{\widetilde{#1}}

\def\xssp{\hspace{1pt}}
\def\ssp{\hspace{3pt}}
\def\msp{\hspace{5pt}}
\def\lsp{\hspace{12pt}}

\newcommand{\cA}{\mathcal{A}}
\newcommand{\cB}{\mathcal{B}}
\newcommand{\cC}{\mathcal{C}}
\newcommand{\cD}{\mathcal{D}}
\newcommand{\cE}{\mathcal{E}}
\newcommand{\cF}{\mathcal{F}}
\newcommand{\cG}{\mathcal{G}}
\newcommand{\cH}{\mathcal{H}}
\newcommand{\cI}{\mathcal{I}}
\newcommand{\cJ}{\mathcal{J}}
\newcommand{\cK}{\mathcal{K}}
\newcommand{\cL}{\mathcal{L}}
\newcommand{\cM}{\mathcal{M}}
\newcommand{\cN}{\mathcal{N}}
\newcommand{\cO}{\mathcal{O}}
\newcommand{\cP}{\mathcal{P}}
\newcommand{\cQ}{\mathcal{Q}}
\newcommand{\cR}{\mathcal{R}}
\newcommand{\cS}{\mathcal{S}}
\newcommand{\cT}{\mathcal{T}}
\newcommand{\cU}{\mathcal{U}}
\newcommand{\cV}{\mathcal{V}}
\newcommand{\cW}{\mathcal{W}}
\newcommand{\cX}{\mathcal{X}}
\newcommand{\cY}{\mathcal{Y}}
\newcommand{\cZ}{\mathcal{Z}}

\newcommand{\vA}{\mathbf{A}}
\newcommand{\vB}{\mathbf{B}}
\newcommand{\vC}{\mathbf{C}}
\newcommand{\vD}{\mathbf{D}}
\newcommand{\vE}{\mathbf{E}}
\newcommand{\vF}{\mathbf{F}}
\newcommand{\vG}{\mathbf{G}}
\newcommand{\vH}{\mathbf{H}}
\newcommand{\vI}{\mathbf{I}}
\newcommand{\vJ}{\mathbf{J}}
\newcommand{\vK}{\mathbf{K}}
\newcommand{\vL}{\mathbf{L}}
\newcommand{\vM}{\mathbf{M}}
\newcommand{\vN}{\mathbf{N}}
\newcommand{\vO}{\mathbf{O}}
\newcommand{\vP}{\mathbf{P}}
\newcommand{\vQ}{\mathbf{Q}}
\newcommand{\vR}{\mathbf{R}}
\newcommand{\vS}{\mathbf{S}}
\newcommand{\vT}{\mathbf{T}}
\newcommand{\vU}{\mathbf{U}}
\newcommand{\vV}{\mathbf{V}}
\newcommand{\vW}{\mathbf{W}}
\newcommand{\vX}{\mathbf{X}}
\newcommand{\vY}{\mathbf{Y}}
\newcommand{\vZ}{\mathbf{Z}}

\newcommand{\va}{\mathbf{a}}
\newcommand{\vb}{\mathbf{b}}
\newcommand{\vc}{\mathbf{c}}
\newcommand{\vd}{\mathbf{d}}
\newcommand{\ve}{\mathbf{e}}
\newcommand{\vf}{\mathbf{f}}
\newcommand{\vg}{\mathbf{g}}
\newcommand{\vh}{\mathbf{h}}
\newcommand{\vi}{\mathbf{i}}
\newcommand{\vj}{\mathbf{j}}
\newcommand{\vk}{\mathbf{k}}
\newcommand{\vl}{\mathbf{l}}
\newcommand{\vm}{\mathbf{m}}
\newcommand{\vn}{\mathbf{n}}
\newcommand{\vo}{\mathbf{o}}
\newcommand{\vp}{\mathbf{p}}
\newcommand{\vq}{\mathbf{q}}
\newcommand{\vr}{\mathbf{r}}
\newcommand{\Vs}{\mathbf{s}}
\newcommand{\vt}{\mathbf{t}}
\newcommand{\vu}{\mathbf{u}}
\newcommand{\vv}{\mathbf{v}}
\newcommand{\vw}{\mathbf{w}}
\newcommand{\vx}{\mathbf{x}}
\newcommand{\vy}{\mathbf{y}}
\newcommand{\vz}{\mathbf{z}}

\newcommand{\vone}{\mathbf{1}}
\newcommand{\vzero}{\mathbf{0}}

\newcommand{\valpha}{{\boldsymbol{\alpha}}}
\newcommand{\vbeta}{{\boldsymbol{\beta}}}
\newcommand{\vgamma}{{\boldsymbol{\gamma}}}
\newcommand{\vdelta}{{\boldsymbol{\delta}}}
\newcommand{\vepsilon}{{\boldsymbol{\epsilon}}}
\newcommand{\vzeta}{{\boldsymbol{\zeta}}}
\newcommand{\veta}{{\boldsymbol{\eta}}}
\newcommand{\vtheta}{{\boldsymbol{\theta}}}
\newcommand{\viota}{{\boldsymbol{\iota}}}
\newcommand{\vkappa}{{\boldsymbol{\kappa}}}
\newcommand{\vlambda}{{\boldsymbol{\lambda}}}
\newcommand{\vmu}{{\boldsymbol{\mu}}}
\newcommand{\vnu}{{\boldsymbol{\nu}}}
\newcommand{\vxi}{{\boldsymbol{\xi}}}
\newcommand{\vomikron}{{\boldsymbol{\omikron}}}
\newcommand{\vpi}{{\boldsymbol{\pi}}}
\newcommand{\vrho}{{\boldsymbol{\rho}}}
\newcommand{\vsigma}{{\boldsymbol{\sigma}}}
\newcommand{\vtau}{{\boldsymbol{\tau}}}
\newcommand{\vupsilon}{{\boldsymbol{\upsilon}}}
\newcommand{\vphi}{{\boldsymbol{\phi}}}
\newcommand{\vchi}{{\boldsymbol{\chi}}}
\newcommand{\vpsi}{{\boldsymbol{\psi}}}
\newcommand{\vomega}{{\boldsymbol{\omega}}}

\newcommand{\rLambda}{\mathrm{\Lambda}}
\newcommand{\rSigma}{\mathrm{\Sigma}}

\newcommand{\vLambda}{\bm{\rLambda}}
\newcommand{\vSigma}{\bm{\rSigma}}


\makeatletter
\newcommand{\vast}[1]{\bBigg@{#1}}
\makeatother

\makeatletter
\newcommand*\bdot{\mathpalette\bdot@{.7}}
\newcommand*\bdot@[2]{\mathbin{\vcenter{\hbox{\scalebox{#2}{$\m@th#1\bullet$}}}}}
\makeatother

\makeatletter
\DeclareRobustCommand\onedot{\futurelet\@let@token\@onedot}
\def\@onedot{\ifx\@let@token.\else.\null\fi\xspace}

\def\eg{\emph{e.g}\onedot} \def\Eg{\emph{E.g}\onedot}
\def\ie{\emph{i.e}\onedot} \def\Ie{\emph{I.e}\onedot}
\def\cf{\emph{cf}\onedot} \def\Cf{\emph{Cf}\onedot}
\def\etc{\emph{etc}\onedot} \def\vs{\emph{vs}\onedot}
\def\wrt{w.r.t\onedot} \def\dof{d.o.f\onedot} \def\aka{a.k.a\onedot}
\def\etal{\emph{et al}\onedot}
\makeatother

\newcommand\amirh[1]{\textcolor{cyan}{AmirH: #1}} %
\newcommand\amirg[1]{\textcolor{red}{AmirG: #1}} %
\newcommand\guillaume[1]{\textcolor{orange}{Guillaume: #1}} %
\newcommand\jens[1]{\textcolor{magenta}{Jens: #1}} %
\newcommand\noor[1]{\textcolor{blue}{Noor: #1}} %
\newcommand\risheek[1]{\textcolor{green}{Risheek: #1}} %

\newcommand{\unet}{\bm{\epsilon}}
\newcommand{\unetH}{\bm{\epsilon}_H}
\newcommand{\unetHin}{\bm{\epsilon}_H^{in}}
\newcommand{\unetHout}{\bm{\epsilon}_H^{out}}
\newcommand{\unetL}{\bm{\epsilon}_L}
\newcommand{\repin}[1]{\bm{r}_{#1}^{in}}
\newcommand{\repout}[1]{\bm{r}_{#1}^{out}}
\newcommand{\repoutpred}[1]{\hat{\bm{r}}_{#1}^{out}}
\newcommand{\adap}{\bm{\phi}_\theta}

\newcommand\method{Clockwork}

\maketitle

\begin{abstract}

This work aims to improve the efficiency of text-to-image diffusion models. While diffusion models use computationally expensive UNet-based denoising operations in every generation step, we identify that not all operations are equally relevant for the final output quality. In particular, we observe that UNet layers operating on high-res feature maps are relatively sensitive to small perturbations. In contrast, low-res feature maps influence the semantic layout of the final image and can often be perturbed with no noticeable change in the output. Based on this observation, we propose \emph{\method\ Diffusion}, a method that periodically reuses computation from preceding denoising steps to approximate low-res feature maps at one or more subsequent steps. For multiple baselines, and for both text-to-image generation and image editing, we demonstrate that \emph{\method} leads to comparable or improved perceptual scores with drastically reduced computational complexity.
As an example, for Stable Diffusion v1.5 with 8 DPM++ steps we save $32\%$ of FLOPs with negligible FID and CLIP change. We release code at \url{https://github.com/Qualcomm-AI-research/clockwork-diffusion}

\end{abstract}

\begin{figure}[t]
    \centering
    \includegraphics[trim={0, 0.2cm, 0, 0}, width=\linewidth]{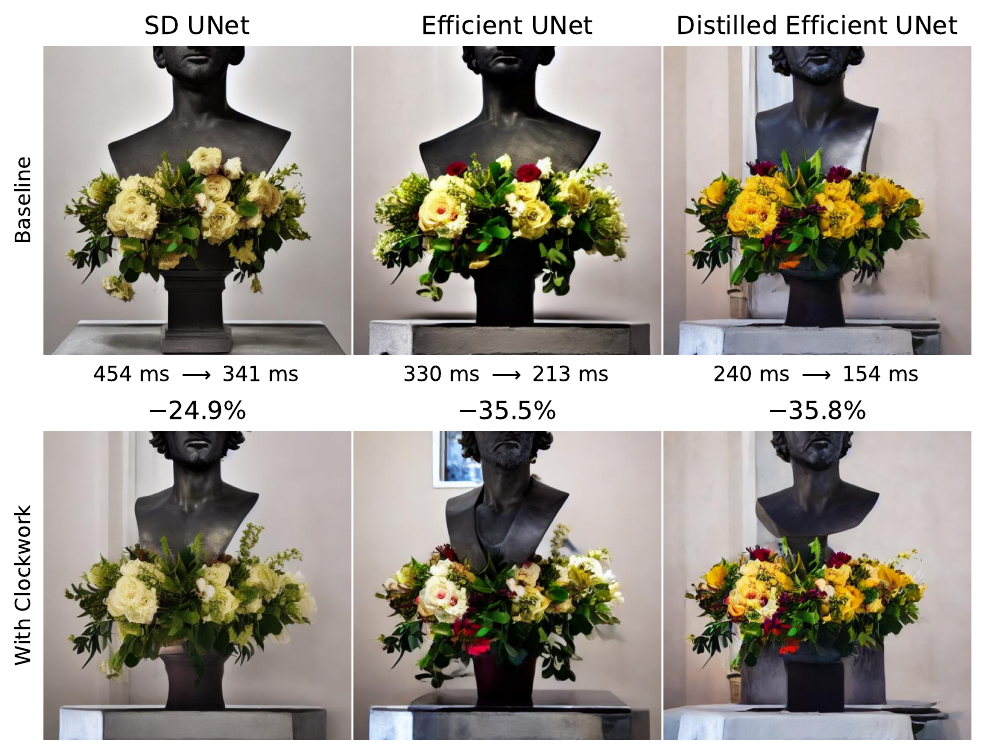}
    \caption{Time savings with \method, for different baselines. All pairs have roughly constant FID (computed on MS-COCO 2017 5K validation set), using 8 sampling steps (DPM++). \method\ can be applied on top of standard models as well as heavily optimized ones. Timings computed on NVIDIA\textsuperscript{\textregistered} RTX\textsuperscript{\textregistered} 3080 at batch size 1 (for distilled model) or 2 (for classifier-free guidance). Prompt: ``the bust of a man's head is next to a vase of flowers''.}
    \vspace{-0.5em}
    \label{fig:fig1}
\end{figure}

\section{Introduction}

Diffusion Probabilistic Models (DPM), or Diffusion Models for short, have become one of the most popular approaches for text-to-image generation\cite{rombach2022high,saharia2022photorealistic}. Compared to Generative Adversarial Networks (GANs), they allow for diverse synthesized outputs and high perceptual quality \cite{dhariwal2021diffusion}, while offering a relatively stable training paradigm \cite{ho2020denoising} and high controllability.

One of the main drawbacks of diffusion models is that they are comparatively slow, involving repeated operation of computationally expensive UNet models \cite{ronnerberger2015unet}. As a result, a lot of current research focuses on improving their efficiency, mainly through two different mechanisms. First, some works seek to \emph{reduce the overall number of sampling steps}, either by introducing more advanced samplers \cite{song2020denoising,lu2022dpm,lu2022dpm++} or by performing so-called step distillation \cite{salimans2021progressive,meng2023distillation}. Second, some works \emph{reduce the required computation per step} \eg, through classifier-free guidance distillation \cite{ho2022classifierfree,meng2023distillation}, architecture search~\cite{li2023snapfusion}, or with model distillation~\cite{kim2023architectural}.

Our work can be viewed as a combination of these two axes. We begin with the observation that lower-resolution representations within diffusion UNets (\ie those further from input and output) are not only influencing the semantic layout more than smaller details \cite{deja2022analyzing,tumanyan2023pnp,si2023freeu}, they are also more resilient to perturbations and thus more amenable to distillation into a smaller model. Hence, we propose to perform model distillation on the lower-resolution parts of the UNet by reusing their representations from previous sampling steps. To achieve this we make several contributions: \textbf{1)} By approximating internal UNet representations with those from previous sampling steps, we are effectively performing a combination of model- and step distillation, which we term \emph{model-step distillation}. \textbf{2)} We show how to design a lightweight adaptor architecture to maximize compute savings, and even show performance improvements by simply caching representations in some cases. \textbf{3)} We show that it is crucial to alternate approximation steps with full UNet passes, which is why we call our method \emph{\method\ Diffusion}. \textbf{4)} We propose a way to train our approach without access to an underlying image dataset, and in less than 24h on a single NVIDIA\textsuperscript{\textregistered} Tesla\textsuperscript{\textregistered} V100 GPU.

We apply \method\ to both text-to-image generation (MS-COCO \cite{lin2014microsoft}) and image editing (ImageNet-R-TI2I \cite{tumanyan2023pnp}), consistently demonstrating savings in FLOPs as well as latency on both GPU and edge device, while maintaining comparable FID and CLIP score. \method\ is complementary to other optimizations like step and guidance distillation~\cite{salimans2021progressive,meng2023distillation} or efficient samplers: we show savings even on an optimized and DPM++ distilled Stable Diffusion model~\cite{rombach2022high,lu2022dpm++}, as can be visualized in \cref{fig:fig1}.

\section{Related work}

\paragraph{Faster solvers.}
Diffusion sampling is equivalent to integration of an ODE or SDE~\cite{song2020scorebased}. As a result, many works attempt to perform integration with as few steps as possible, often borrowing from existing literature on numerical integration. DDIM~\cite{song2021denoising} introduced deterministic sampling, drastically improving over the original DDPM \cite{ho2020denoising}. Subsequently, works have experimented with multistep~\cite{liu2022pndm}, higher-order solvers~\cite{jolicoeurmartineau2021gotta,karras2022elucidating,dockhorn2022genie}, predictor-corrector methods~\cite{zhang2022gddim,zhao2023unipc}, or combinations thereof. DPM++~\cite{lu2022dpm++,lu2022dpm} stands out as one of the fastest solvers, leveraging exponential integration, and we conduct most of our experiments with it. However, in our ablation studies in the Appendix-\cref{tab:clockwork_schedulers}, we show that the benefit of \method\ is largely independent of the choice of solver.

\paragraph{Step Distillation}
 starts with a trained teacher model, and then trains a student to mirror the output of multiple teacher model steps \cite{luhman2021knowledge,salimans2021progressive}. It has been extended to guided diffusion models ~\cite{meng2023distillation,li2023snapfusion}, where Meng \etal \cite{meng2023distillation} first distill unconditional and conditional model passes into one and then do step distillation following\cite{salimans2021progressive}. Berthelot \etal \cite{berthelot2023tract} introduce a multi-phase distillation technique similar to Salimans and Ho \cite{salimans2021progressive}, but generalize the concept of distilling to a student model with fewer iterations beyond a factor of two. Other approaches do not distill students to take several steps simultaneously, but instead aim to distill straighter sampling trajectories, which then admit larger step sizes for integration\cite{song2023consistency,liu2022flow,liu2023instaflow}. In particular, InstaFlow \cite{liu2023instaflow} shows impressive results with single-step generation.

Our approach incorporates ideas from step distillation wherein internal UNet representations from previous steps are used to approximate the representations at the same level for the current step. At the same time, it is largely orthogonal and can be combined with the above. We demonstrate savings on an optimized Stable Diffusion model with step and guidance distillation.

\paragraph{Efficient Architectures.}
To reduce the architecture complexity of UNet, \emph{model or knowledge distillation} techniques have been adopted either at output level or feature level~\cite{kim2023architectural,li2023snapfusion,dockhorn2023distilling}. Model pruning~\cite{choi2023squeezing, li2023snapfusion} and 
model quantization~\cite{shang2023post,he2023ptqd,pandey2023softmax} have also been explored to accelerate inference at lower precision while retaining quality. Another direction has been to optimize kernels for faster on-device inference \cite{chen2023speed}, but such solutions are hardware dependent.

Our work can be considered as model distillation, as we replace parts of the UNet with more lightweight components.
But unlike traditional model distillation, we only replace the full UNet for \emph{some steps in the trajectory}. Additionally, we provide our lightweight adaptors outputs from previous steps, making it closer to step distillation.

\begin{figure*}[t]
    \centering
    \includegraphics[trim={0, 0.6cm, 0, 0},width=0.85\textwidth]{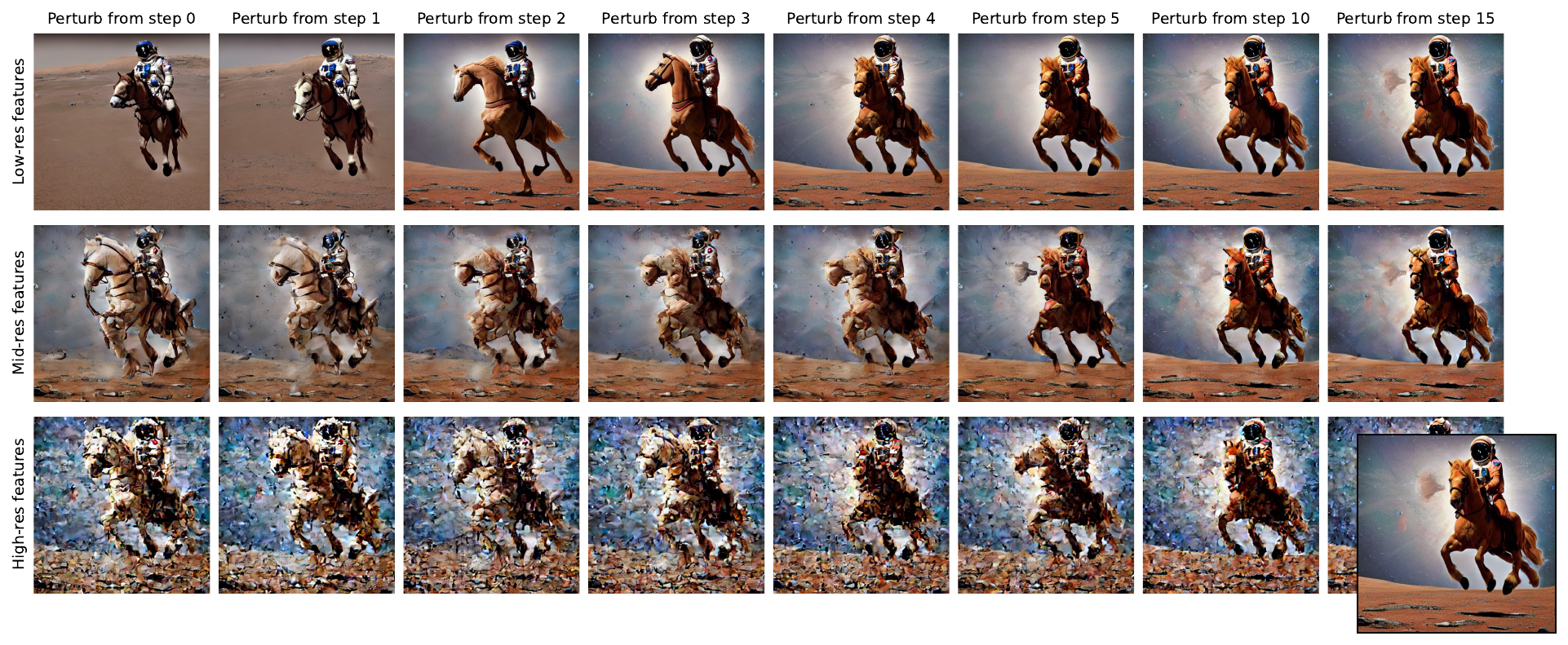}
    \caption{Perturbing Stable Diffusion v1.5 UNet representations (outputs of the three upsampling layers), starting from different sampling steps (20 DPM++ steps total, note the reference image as inset in lower-right). Perturbing low-resolution features after only a small number of steps has a comparatively small impact on the final output, whereas perturbation of higher-res features results in high-frequency artifacts. Prompt: "image of an astronaut riding a horse on mars."}
    \vspace{-1em}
    \label{fig:perturbation_qualitative}
\end{figure*}

\section{Analysis of perturbation robustness}
\label{sec:analysis}

Our method design takes root in the observation that lower-resolution features in diffusion UNets are robust to perturbations, as measured by the change in the final output. This section provides a qualitative analysis of this behaviour.

During diffusion sampling, earlier steps contribute more to the semantic layout of the image, while later steps are more related to high-frequency details \cite{deja2022analyzing,si2023freeu}. Likewise, lower-res UNet representations contribute more to the semantic layout, while higher-res features and skip connections carry high-frequency content \cite{tumanyan2023pnp,si2023freeu}. This can be leveraged to perform image editing at a desired level of detail by performing DDIM inversion \cite{song2020scorebased} and storing feature and attention maps to reuse during generation \cite{tumanyan2023pnp}. We extend this by finding that the lower-res representations, which contribute more to the semantic layout, are also more robust to perturbations. This makes them more amenable to distillation.

For our illustrative example, we choose random Gaussian noise to perturb feature maps. In particular, we mix a given representation with a random noise sample in a way that keeps activation statistics roughly constant. We assume a feature map to be normal $\bm{f}\sim\cN(\mu_f,\sigma_f^2)$, and draw a random sample $\bm{z}\sim\cN(0,\sigma_f^2)$. We then update the feature map with:

\begin{equation}
    \bm{f}\leftarrow\mu_f+\sqrt{\alpha}\cdot(\bm{f}-\mu_f)+\sqrt{1-\alpha}\cdot\bm{z}
\end{equation}

On average, this will leave the distribution unchanged. We set $\alpha=0.3$ to make the noise the dominant signal.

In \cref{fig:perturbation_qualitative} we perform such perturbations on the outputs of the three upsampling layers of the Stable Diffusion v1.5 UNet \cite{rombach2022high}. Perturbation starts after a varying number of unperturbed steps and the final output is shown for each case.
After only a small number of steps the lowest-resolution features can be perturbed without a noticeable change in the final output, whereas higher-res features are affected for longer along the trajectory. Moreover, early perturbations in lower-res layers mostly result in semantic changes, confirming findings from other works \cite{deja2022analyzing,si2023freeu}.
Implementation details and additional analyses for other layers are provided in \cref{sec:appendix:analysis}.

Motivated by these findings, we propose to approximate lower-res UNet representations using more computationally lightweight functions, and in turn reuse information from previous sampling steps, effectively combining model and step distillation. However, we make another crucial and non-trivial contribution. \cref{fig:perturbation_qualitative} might suggest that one should approximate all representations after a certain sampling step. We instead find that it is beneficial to alternate approximation steps and full UNet passes to avoid accumulating errors. This makes our approach similar to others that run model parts with different temporal granularity \cite{koutnk2014clockwork,shelhamer2016clockwork}, and we consequently name it \emph{\method\ Diffusion}.


\begin{figure*}[t]
    \centering
    \includegraphics[width=0.95\textwidth]{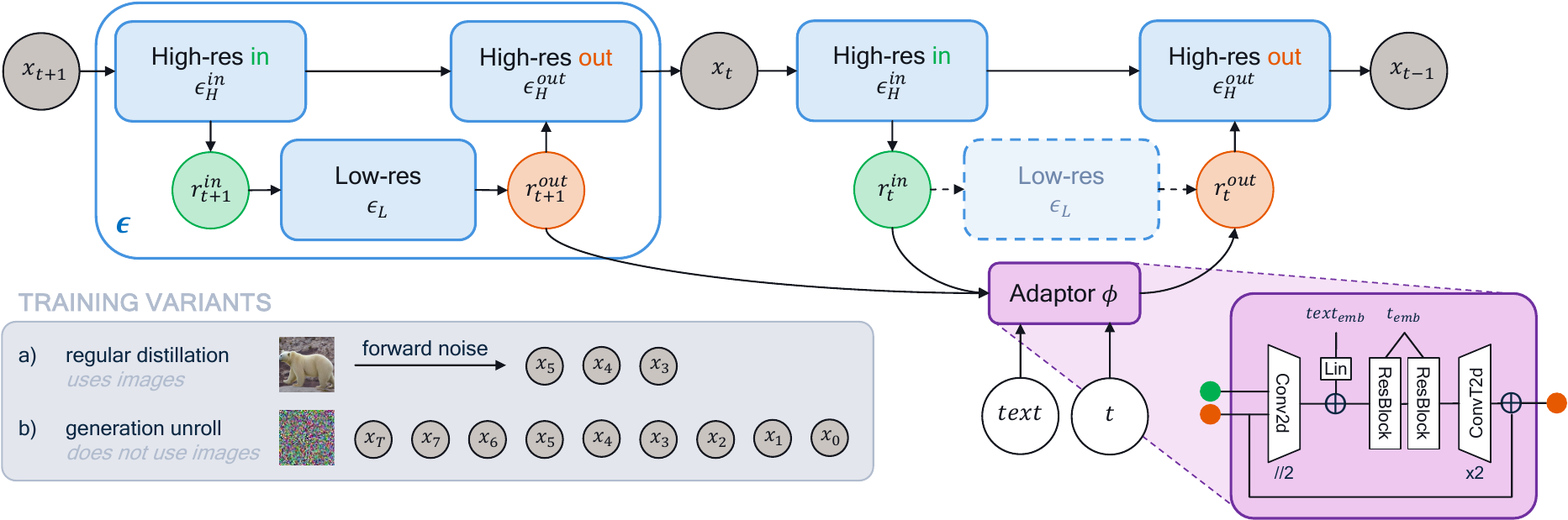}
    \caption{Schematic view of \emph{\method}. It can be thought of as a combination of model distillation and step distillation. We replace the lower-resolution parts of the UNet $\unet$ with a more lightweight adaptor, and at the same time give it access to features from the previous sampling step. Contrary to common step distillation, which constructs latents by forward noising images, we train with sampling trajectories unrolled from pure noise. Other modules are conditioned on text and time embeddings (omitted for readability). The gray panel illustrates the difference between regular distillation and our proposed training with unrolled trajectories.}    
    \label{fig:architecture}
\end{figure*}

\section{Clockwork Diffusion}
\label{sec:clockwork}
Diffusion sampling involves iteratively applying a learned denoising function $\unet_\theta(\cdot)$, or an equivalent reparametrization, to denoise a noisy sample $\vx_t$ into a less noisy sample $\vx_{t-1}$ at each iteration $t$, starting from a sample from Gaussian noise at $t=T$ towards a final generation at $t=0$~\cite{sohldickstein2015deep,ho2020denoising}.

As is illustrated in \cref{fig:architecture}, the noise prediction function $\unet$ (we omit the parameters $\theta$ for clarity) is most commonly implemented as a UNet, which can be decomposed into low- and high-resolution denoising functions $\unetL$ and $\unetH$ respectively. $\unetH$ further consists of an input module $\unetHin$ and an output module $\unetHout$, where $\unetHin$ receives the diffusion latent $\vx_t$ and $\unetHout$ predicts the next latent $\vx_{t-1}$ (usually not directly, but by estimating its corresponding noise vector or denoised sample).
The low-resolution path $\unetL$ receives a lower-resolution internal representation $\repin{t}$ from $\unetHin$ and predicts another internal representation $\repout{t}$ that is used by $\unetHout$.
We provide a detailed view of the architecture and how to separate it in the \cref{sec:appendix:arch_details}.

The basis of \emph{\method\ Diffusion} is the realization that the outputs of $\unetL$ are relatively robust to perturbations --- as demonstrated in \cref{sec:analysis} --- and that it should be possible to approximate them with more computationally lightweight functions if we reuse information from previous sampling steps. The latter part differentiates it from regular model distillation \cite{kim2023architectural,dockhorn2023distilling}. Overall, there are 4 key contributions that are necessary for optimal performance: \textbf{a)} joint model and step distillation, \textbf{b)} efficient adaptor design, \textbf{c)} \emph{\method} scheduling, and \textbf{d)} training with unrolled sampling trajectories. We describe each below.

\subsection{Model-step distillation}
\label{sub:model_step_distillation}

\emph{Model distillation} is a well-established concept where a smaller student model is trained to replicate the output of a larger teacher model, operating on the same input. \emph{Step distillation} is a common way to speed up sampling for diffusion models, where a student is trained to replace e.g. two teacher model passes. Here the input/output change, but the model architecture is usually kept the same. We propose to combine the two, replacing part of the diffusion UNet with a more lightweight adaptor, but in turn giving it access to outputs from previous sampling steps (as shown in \cref{fig:architecture}). We term this procedure \emph{model-step distillation}.

In its simplest form, an adaptor $\adap$ is an identity mapping that naively copies a representation $\bm{r}^{out}$ from step $t+1$ to $t$. This works relatively well when the number of sampling steps is high, as for example in our image editing experiments in \cref{sec:experiments_image_to_image}. For a more effective approximation in the low step regime, we rely on a parametric function $\adap$ with additional inputs: $\repoutpred{t} = \adap\left(\repin{t},\repout{t+1},\bm{t}_{emb},\bm{text}_{emb}\right)$, which we describe as follows.

\subsection{Efficient adaptor architecture}
\label{sub:efficient_adaptor}

The design of our adaptor is chosen to minimize heavy compute operations. It uses no attention, and is instead comprised of a strided convolutional layer resulting in two times spatial downsampling, followed by addition of a linear projection of the prompt embedding, two ResNet blocks with additive conditioning on $\bm{t}$, and a final transposed convolution to go back to the original resolution. We further introduce a residual connection from input to output. The adaptor architecture is shown in \cref{fig:architecture}, and we provide more details in \cref{sec:appendix:arch_details}. We ablate several architecture choices in \cref{sec:experiments_ablation}. The inputs to the adaptor are listed below. 

\paragraph{Input representation $\repin{t}$} is the representation obtained from the high-res input module $\unetHin$ at the current step, as shown in \cref{fig:architecture}. It is concatenated with the next input.
\paragraph{Output representation $\repout{t+1}$} is the equivalent representation from the previous sampling step that the adaptor tries to approximate for the current step. The high-res output module predicts the next diffusion latent from it. By conditioning on $\repout{t+1}$, our approach depends on the sampler and step width (similar to step distillation).
\paragraph{Time embedding $\bm{t}_{emb}$} is an additional input to the adaptor to make it conditional on the diffusion step $t$, instead of training separate adaptor models for each step. For this purpose we rely on the standard ResBlocks with time step embeddings, as in Rombach \etal~\cite{rombach2022high}.
\paragraph{Prompt embedding $\bm{text}_{emb}$} is an additional input to the adaptor to make it conditional on the generation prompt. We rely on the \emph{pooled} CLIP embedding \cite{radford2021learning} of the prompt, extracted using OpenCLIP's ViT-g/14 \cite{ilharco2021openclip}, instead of the sequence to reduce the complexity.

\subsection{Clockwork scheduling}

Instead of just replacing $\unetL$ with an adaptor $\adap$ entirely, we avoid accumulating errors during sampling by alternating lightweight adaptor steps with full UNet passes, which is the inspiration for our method's name, following \cite{koutnk2014clockwork,shelhamer2016clockwork}. Specifically, we switch between $\unetL$ and $\adap$ based on a predefined clock schedule $\cC(t) \in \{0, 1\}$ as follows:

\begin{equation*}
\label{eq:z_t}
    \repoutpred{t} =
    \begin{cases}
        \unetL\left(\repin{t},\bm{t}_{emb},\bm{text}_{emb}\right), & \cC(t) = 0 \\
        \adap\left(\repin{t},\repout{t+1},\bm{t}_{emb},\bm{text}_{emb}\right), & \cC(t) = 1
    \end{cases}
\end{equation*}

where $\bm{t}$ and $\bm{c}$ are time step and prompt embeddings, respectively. $\cC(t)$ can generally be an arbitrary schedule of switches between $\unetL$ and $\adap$, but we find that interleaving them at a fixed rate offers a good tradeoff between performance and simplicity. Because we conduct our experiments mostly in the low-step regime with $\leq8$ steps, we simply alternate between adaptor and full UNet in consecutive steps (\ie a \emph{clock} of 2) unless otherwise specified. For sampling with more steps it is possible to use more consecutive adaptor passes, as we show in \Cref{sec:appendix:pnp:quantitative} for the text-guided image editing case. For the rest of the paper, we simply use the terminology \emph{a clock of $N$}, which means every $N$ steps, a full UNet pass will be evaluated, all other steps use the adaptor.

\subsection{Distillation with unrolled trajectories}
\label{sub:adaptor_training}

We seek to train an adaptor that predicts an internal UNet representation, based on the same representation from the previous sampling step as well as further inputs. Formally, we minimize the following loss:

\begin{equation}
    \cL = \mathop{{}\expect}_t\left[\norm{\repout{t}-\adap\left(\repin{t},\repout{t+1},\bm{t}_{emb},\bm{text}_{emb}\right)}_2\right]
\end{equation}

A common choice is to stochastically approximate the expectation over update steps, \ie just sample $t$ randomly at each training step. Most step distillation approaches \cite{salimans2021progressive,meng2023distillation} then construct $\vx_t$ from an image $\vx_0$ via the diffusion forward process, and perform two UNet passes of a teacher model to obtain all components required for the loss. Instead of this, we start from a random noise sample and unroll a full sampling trajectory $\{\vx_T,\ldots,\vx_0\}$ with the teacher model, then use each step as a separate training signal for the adaptor. This is illustrated in \cref{fig:architecture}. We construct a dataset of unrolled sampling trajectories for each epoch, which can be efficiently parallelized using larger batch sizes. We compare our unrolled training with the conventional approach in \cref{sec:experiments_ablation}.

Overall training can be done in less than a day on a single NVIDIA\textsuperscript{\textregistered} Tesla\textsuperscript{\textregistered} V100 GPU. As an added benefit, this training scheme does not require access to an image dataset and only relies on captions. We provide more details in \cref{sec:experiments} and include training pseudo-code in Appendix-\cref{alg:unrolled_training_alg}.

\section{Experiments}
\label{sec:experiments}
We evaluate the effectiveness of \method\ on two tasks: text-guided image generation in \cref{sec:experiments_text_to_image} and text-guided image editing in \cref{sec:experiments_image_to_image}. Additionally, we provide several ablation experiments in ~\cref{sec:experiments_ablation}.

\subsection{Experimental setup}
\label{sec:experiments_setup}
\paragraph{Datasets and metrics}
We evaluate our text-guided image generation experiments by following common practices~\cite{rombach2022high, li2023snapfusion, meng2023distillation} on two public benchmarks: MS-COCO 2017 (5K captions), and MS-COCO 2014~\cite{lin2014microsoft} (30K captions) validation sets. We use each caption to generate an image and rely on the CLIP score from a OpenCLIP ViT-g/14 model~\cite{ilharco2021openclip} to evaluate the alignment between captions and generated images. We also rely on Fréchet Inception Distance (FID) \cite{heusel2017fid} to estimate perceptual quality. For MS-COCO 2014, the images are resized to $256\times256$ before computing the FID as in Kim \etal~\cite{kim2023architectural}.
We evaluate our text-guided image editing experiments on the ImageNet-R-TI2I~\cite{tumanyan2023pnp} dataset that includes various renderings of ImageNet-R~\cite{hendrycks2021many} object classes. Following~\cite{tumanyan2023pnp}, we use 3 high-quality images from 10 different classes and 5 prompt templates to generate 150 image-text pairs for evaluation. In addition to the CLIP score, we measure the DINO self-similarity distance as introduced in Splice~\cite{tumanyan2022splice} to measure the structural similarity between the source and target images.

To measure the computational cost of the different methods, we report the time spent on latent generation, which we call \emph{latency} for short, as it represents the majority of the total processing time. This measures the cost spent on UNet forward passes during the generation --- and inversion in case of image editing --- but ignores the fixed cost of text encoding and VAE decoding. Along with latencies we report the number of floating point operations (FLOPs). We measure latency using PyTorch's benchmark utilities on a single NVIDIA\textsuperscript{\textregistered} RTX\textsuperscript{\textregistered} 3080 GPU, and use the DeepSpeed~\cite{rasley2020deepspeed} library to estimate the FLOP count.
Finally, to verify the efficiency of \method\ on low-power devices, we measure its inference time on a Samsung Galaxy S23 device. It uses a Qualcomm ``Snapdragon\textsuperscript{\textregistered} 8 Gen. 2 Mobile Platform'' with a Qualcomm\textsuperscript{\textregistered} Hexagon\textsuperscript{\tiny{TM}} processor

\paragraph{Diffusion models}
We evaluate the effectiveness of \method\ on three latent diffusion models with varying computational costs:
\emph{i)} \textbf{\textit{SD UNet}}, the standard UNet from Stable Diffusion v1.5~\cite{rombach2022high}.
\emph{ii)} \textbf{\textit{Efficient UNet}}, which, inspired by Li \etal~\cite{li2023snapfusion}, removes the costly transformer blocks, including self-attention and cross-attention operations, from the highest resolution layer of SD UNet.
\emph{iii)} \textbf{\textit{Distilled Efficient UNet}}, which further accelerates Efficient UNet by implementing progressive step distillation~\cite{salimans2021progressive} and classifier-free guidance distillation~\cite{meng2023distillation}. Since there is no open source implementation~\cite{li2023snapfusion, salimans2021progressive, meng2023distillation} available, we rely on our replication as specified in the supplementary materials.
In all experiments we use the DPM++~\cite{lu2022dpm++} multi-step scheduler due to its superiority in the low number of sampling steps regime, which is a key focus of our paper. An exception is the text-guided image editing experiment where we use the DDIM scheduler as in Plug-and-Play~\cite{tumanyan2023pnp}.

\paragraph{Implementation details}
We train \method\ using a ResNet-based adaptor (as shown in \cref{fig:architecture}) for a specific number of generation steps $T$ and with a clock of 2, as described in \cref{sub:model_step_distillation}, on 50K random captions from the LAION-5B dataset~\cite{schuhmann2022laion}. The training involves $120$ epochs using the Adam optimizer~\cite{kingma2017adam} with a batch size of $16$ and learning rate of $0.0001$. Thanks to its parameter efficiency each training takes less than one day on a single NVIDIA\textsuperscript{\textregistered} Tesla\textsuperscript{\textregistered} V100 GPU.

\begin{figure}
    \centering
    \includegraphics[width=0.95\linewidth]{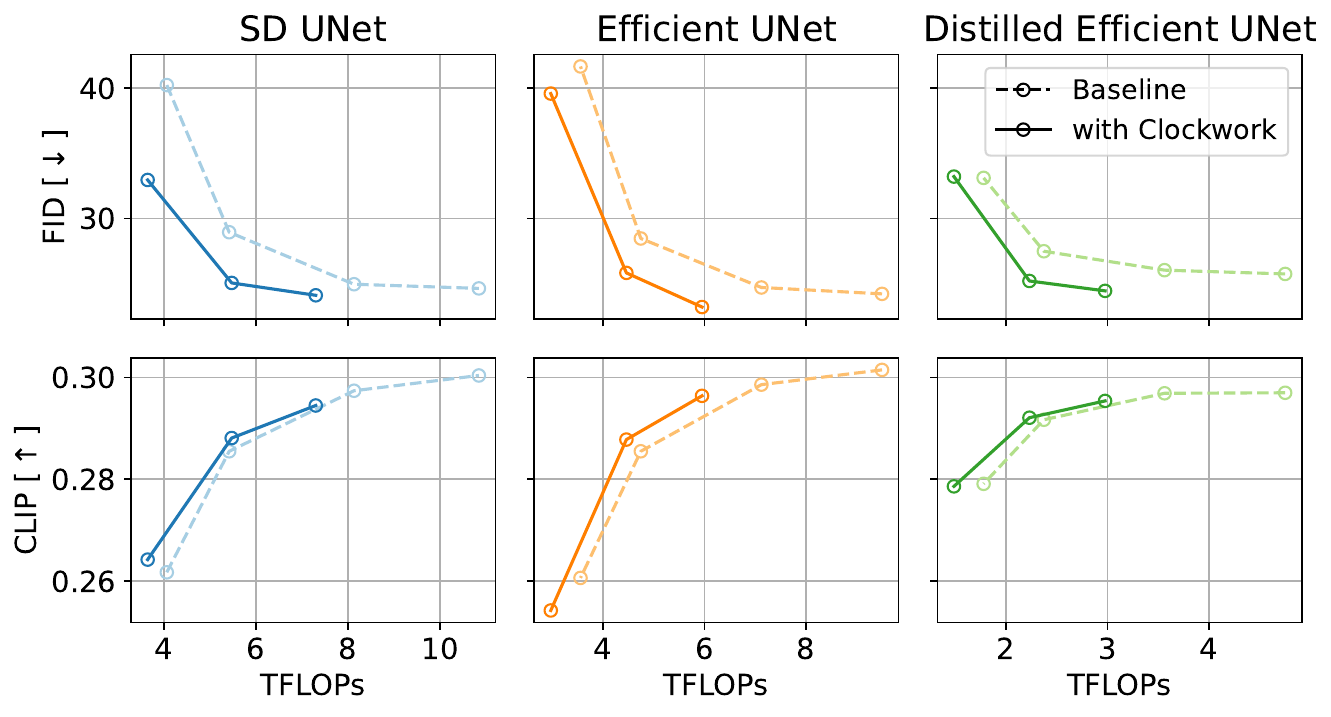}
    \caption{\method\ improves text-to-image generation efficiency consistently over various diffusion models. Models are evaluated on $512\times512$ \textbf{MS-COCO 2017-5K} validation set.
    }
    \label{fig:sota_t2i_plots}
\end{figure}

\subsection{Text-guided image generation}
\label{sec:experiments_text_to_image}
\begin{figure}
    \centering
    \includegraphics[width=\linewidth]{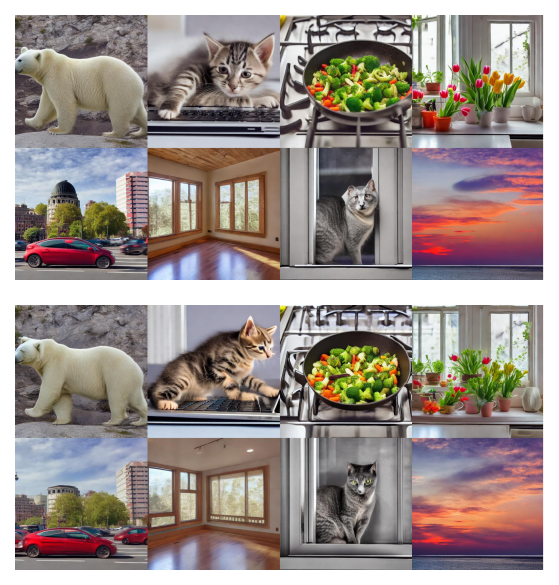}
    \caption{Text guided generations by SD UNet without (top) and with (bottom) \method\ at 8 sampling steps (DPM++). \method\ reduces FLOPs by $32\%$ at a similar generation quality. Prompts given in \cref{sec:appendix:t2i_examples}.}
    \label{fig:t2i_examples_2}
\end{figure}

We evaluate the effectiveness of \method\ in accelerating text-guided image generation for three different diffusion models as specified in~\cref{sec:experiments_setup}. For each model, we measure the generation quality and computational cost using $8$, $6$ and $4$ steps with and without clockwork, as shown in~\cref{fig:sota_t2i_plots}. For the baselines (dashed lines) we also include a point with $3$ sampling steps as a reference. Our results demonstrate that applying \method\ for each model results in a high reduction in FLOPs with little changes in generation qualities (solid lines). For example, at 8 sampling steps, \method\ reduces the FLOPs of the distilled Efficient UNet by $38\%$ from $4.7$ TFLOPS to $2.9$ TFLOPS with only a minor degradation in CLIP ($0.6\%$) and improvement in FID ($5\%$). \cref{fig:t2i_examples_2} shows generation examples for Stable Diffusion with and without \method, while \cref{fig:fig1} shows an example for Efficient UNet and its distilled variant. See \cref{sec:appendix:t2i_examples} for more examples.

Our improvement on the distilled Efficient UNet model demonstrates that \method\ is complementary to other acceleration methods and adds savings on top of step distillation~\cite{salimans2021progressive}, classifier-free guidance distillation~\cite{meng2023distillation}, efficient backbones~\cite{li2023snapfusion} and efficient noise schedulers~\cite{lu2022dpm++}. Moreover, \method\ consistently improves the diffusion efficiency at very low sampling steps, which is the critical operating point for most time-constrained real-world applications, \eg image generation on phones.

In \cref{tab:sota_t2i_coco2017} and \cref{tab:sota_t2i_coco2014} we compare \method\ to state-of-the-art methods for efficient diffusion on MS-COCO 2017 and 2014 respectively. The methods include classifier-free guidance distillation by Meng \etal \cite{meng2023distillation}, SnapFusion \cite{li2023snapfusion}, model distillation from BK-SDM \cite{kim2023architectural} and InstaFlow\cite{liu2023instaflow}. For BK-SDM \cite{kim2023architectural} we use models available in the diffusers library \cite{vonplaten2022diffusers} for all measurements. For Meng \etal \cite{meng2023distillation}, SnapFusion \cite{li2023snapfusion} and InstaFlow (1 step) \cite{liu2023instaflow} we report scores from the original papers and implement their architecture to measure latency and FLOPS. In terms of quantitative performance scores, \method\ improves FID and slightly reduces CLIP on both datasets. Efficient UNet + \method\ achieves the best FID out of all methods. 
InstaFlow has lowest FLOPs and latency as they specifically optimize the model for single-step generation, however, in terms of FID and CLIP, \method\ is significantly better.
Compared to SnapFusion, which is optimized and distilled from the same Stable Diffusion model, our Distilled Efficient UNet + \method\ is significantly more compute efficient and faster.

\subsection{Text-guided image editing}
\label{sec:experiments_image_to_image}

We apply our method to a recent text-guided image-to-image (TI2I) translation method called Plug-and-Play (PnP) \cite{tumanyan2023pnp}. The method caches convolutional features and attention maps during source image inversion \cite{song2020scorebased} at certain steps early in the trajectory. These are then injected during the generation using the target prompt at those same steps. This enables semantic meaning of the original image to be preserved, while the self-attention keys and queries allow preserving the guidance structure.

PnP, like many image editing works \cite{kim2022diffusionclip,hertz2022prompttoprompt,parmar2023pix2pixzero}, requires DDIM inversion \cite{song2020scorebased}. Inversion can quickly become the complexity bottleneck, as it is often run for many more steps than the generation. For instance, PnP uses 1000 inversion steps and 50 generation steps.
We focus on evaluating PnP and its \method\ variants on the ImageNet-R-TI2I \emph{real} dataset with SD UNet. Contrary to the rest of the paper, we use the DDIM sampler for these experiments to match PnP's setup. To demonstrate the benefit of \method\ in a training-free setting, we use an identity adaptor with a clock of 2 \emph{both} in inversion and generation. We use the official open-source diffusers \cite{vonplaten2022diffusers} implementation\footnote{\url{https://github.com/MichalGeyer/pnp-diffusers}} of PnP for these experiments, details in \cref{subsec:appendix:pnp:details}.

\begin{table}[t!]
\centering
\resizebox{\columnwidth}{!}{
\def\arraystretch{1.2}%
\begin{tabular}{lccccc} 
\toprule
   \textbf{Model} & \textbf{FID [$\downarrow$]} & \textbf{CLIP [$\uparrow$]} & \textbf{TFLOPs}& \textbf{Latency} \textcolor{gray}{\small {(GPU)}} & \textbf{Latency} \textcolor{gray}{\small {(Phone)}} \\ 
   \midrule
    Meng~\etal \cite{meng2023distillation}              & 26.9 & 0.300 & 6.4 & 320 & -\\ 
    SnapFusion~\cite{li2023snapfusion}                  & 24.20 & 0.300 & 4.0 & 185 & -\\
    BK-SDM-Base \cite{kim2023architectural}             & 29.26 & 0.291 & 8.4 & 348 & -\\ 
    BK-SDM-Small \cite{kim2023architectural}            & 29.48 & 0.272 & 8.2 & 336 & -\\
    BK-SDM-Tiny \cite{kim2023architectural}             & 31.48 & 0.268 & 7.8 & 313 & -\\
    InstaFlow (1 step) \cite{liu2023instaflow}             & 29.30 & 0.283 & \textbf{0.8} & \textbf{40} & - \\

    \midrule
    SD UNet                                            & 24.64 & 0.300 & 10.8 & 454 & 3968\\ 
    \cellcolor{cyan!5} \quad + \method                 & \cellcolor{cyan!5} 24.11 & \cellcolor{cyan!5} 0.295 & \cellcolor{cyan!5} 7.3 ($-32\%$) & \cellcolor{cyan!5} 341 ($-25\%$) & \cellcolor{cyan!5} 3176 ($-20\%$)\\
    Efficient UNet                                     & 24.22 & \textbf{0.302} & 9.5 & 330 & 1960\\ 
    \cellcolor{cyan!5} \quad + \method                 &\cellcolor{cyan!5} \textbf{23.21} & \cellcolor{cyan!5} 0.296 & \cellcolor{cyan!5} 5.9 ($-38\%$)& \cellcolor{cyan!5} 213 ($-36\%$) & \cellcolor{cyan!5} 1196 ($-39\%$)\\
    Distilled Efficient UNet                            & 25.75 & 0.297 & 4.7 & 240 & 980\\
    \cellcolor{cyan!5} \quad + \method               & \cellcolor{cyan!5} 24.45 & \cellcolor{cyan!5} 0.295 & \cellcolor{cyan!5} 2.9 ($-38\%$) & \cellcolor{cyan!5} 154 ($-36\%$) & \cellcolor{cyan!5} \textbf{598} ($-39\%$)\\

\bottomrule
\end{tabular}
}
\caption{Text guided image generation results on $512\times512$ \textbf{MS-COCO 2017-5K} validation set. We compare to state-of-the-art efficient diffusion models, all at $8$ sampling steps (DPM++) except when specified otherwise. Latency measured in ms. 
}
\label{tab:sota_t2i_coco2017}
\vspace{-0.7em}
\end{table}

\begin{figure*}
    \centering
    \includegraphics[trim={0, 0.7cm, 0, 0},width=0.98 \textwidth]{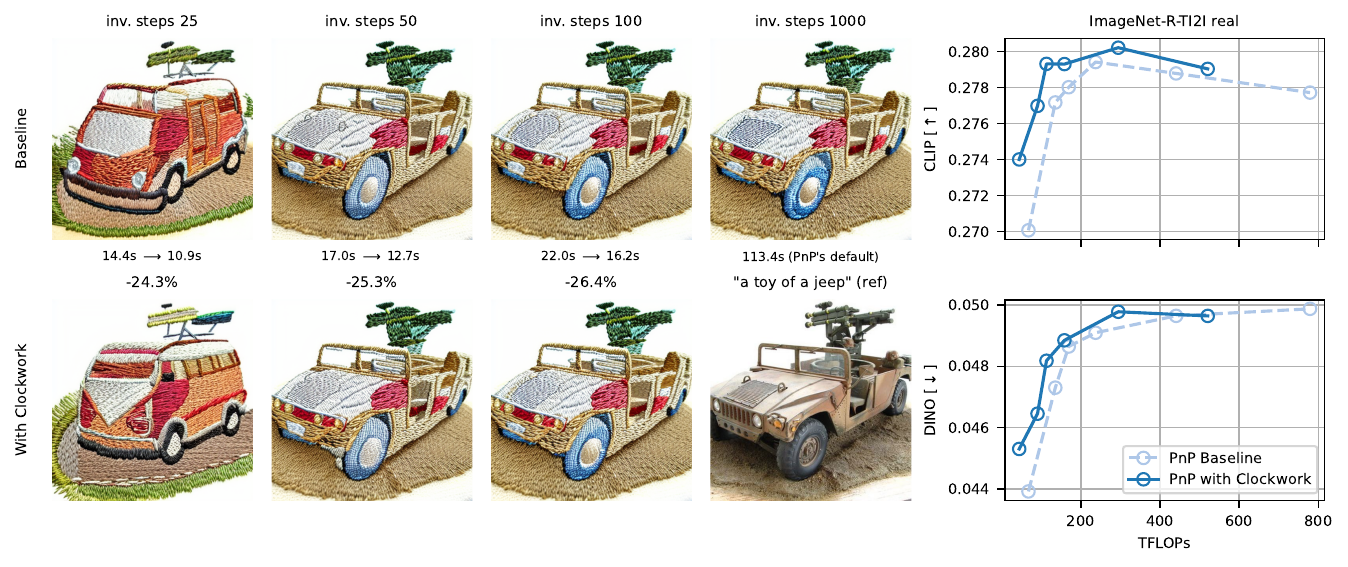}
    \caption{Left: text-guided image editing qualitative results comparing the baseline Plug-and-Play to \method\ with identity adaptor when using the reference image (bottom right) with the target prompt ``an embroidery of a minivan''. Across configurations, applying \method\ enables matching or outperforming the perceptual quality of the baseline Plug-and-Play while reducing latency by a significant margin. Right: \method\ improves the efficiency of text-guided image translation on the ImageNet-R-TI2I real dataset. We evaluate both the baseline and its \method\ variant at different number of DDIM inversion steps: 25, 50, 100, 500 and 1000. The number of DDIM generation steps is fixed to 50 throughout, except for 25 where we use the same number of generation steps as inversion steps.}
    \label{fig:ti2i-pnp-qualitative}
\end{figure*}

In \cref{fig:ti2i-pnp-qualitative} we show qualitative examples of the same text-image pair with and without \method\, for different DDIM inversion steps and generation fixed to 50 steps. For high numbers of inversion steps, \method\ leads to little to no degradation in quality while consistently reducing latency by about $25\%$. At lower numbers of inversions steps, where less features can be extracted (and hence injected at generation), \method\ outputs start diverging from the baseline's, yet in semantically meaningful and perceptually pleasing ways. 

On the right hand side of \cref{fig:ti2i-pnp-qualitative}, we quantitatively show how, for various number of inversion steps, applying \method\ enables saving computation cycles while improving text-image similarity and only slightly degrading structural distance. For PnP's default setting of 1000 inversion steps and 50 generation steps (rightmost point on each curve) \method\ allows saving 33\% of the computational cycles while significantly improving CLIP score, and only slightly degrading DINO self-similarity.

\subsection{Ablation analysis}
\label{sec:experiments_ablation}
In this section we inspect different aspects of Clockwork. For all ablations, we follow the same training procedure explained in~\cref{sec:experiments_setup} and evaluate on the MS-COCO 2017 dataset, with a clock of $2$ and Efficient Unet as backbone. Further ablations, \eg results on different solvers, adaptor input variations are shown in \cref{sec:appendix:ablation}.

\begin{table}[t!]
\centering
\resizebox{.8\columnwidth}{!}{
\def\arraystretch{1.}%
\begin{tabular}{lccc} 
\toprule
   \textbf{Model} & \textbf{FID [$\downarrow$]} & \textbf{CLIP [$\uparrow$]} & \textbf{TFLOPs}\\ 
   \midrule
    SnapFusion~\cite{li2023snapfusion}                  & 14.00 & \textbf{0.300} & 4.0 \\
    BK-SDM-Base \cite{kim2023architectural}             & 17.23 & 0.287 & 8.4\\
    BK-SDM-Small \cite{kim2023architectural}            & 17.72 & 0.268 & 8.2 \\
    BK-SDM-Tiny \cite{kim2023architectural}             & 18.64 & 0.265 & 7.8 \\
    InstaFlow (1 step) \cite{liu2023instaflow}          & 20.00 & - & \textbf{0.8} \\
    \midrule
    SD UNet                                            & 12.77 & 0.296 & 10.8 \\
    \cellcolor{cyan!5} \quad + \method               & \cellcolor{cyan!5} 12.27 & \cellcolor{cyan!5} 0.291 & \cellcolor{cyan!5} 7.3 ($-32\%$)\\
    Efficient UNet                                     & 12.33 & 0.296 & 9.5\\
    \cellcolor{cyan!5} \quad + \method                                  &\cellcolor{cyan!5} \textbf{11.14} & \cellcolor{cyan!5} 0.290 & \cellcolor{cyan!5} 5.9 ($-38\%$)\\
    Distilled Efficient UNet                            & 13.92 & 0.292  & 4.7 \\
    \cellcolor{cyan!5} \quad + \method               & \cellcolor{cyan!5} 12.37 & \cellcolor{cyan!5} 0.291 & \cellcolor{cyan!5}2.9 ($-38\%$) \\
\bottomrule
\end{tabular}
}
\caption{Text guided image generation results on $256\times256$ \textbf{MS-COCO 2014-30K} validation set. We compare to state-of-the-art efficient diffusion models. Except for InstaFlow\cite{liu2023instaflow} all models are evaluated at $8$ sampling steps using the DPM++ scheduler.
}
\label{tab:sota_t2i_coco2014}
\vspace{-0.7em}
\end{table}

\begin{figure}
    \centering
    \includegraphics[width=0.99\linewidth]{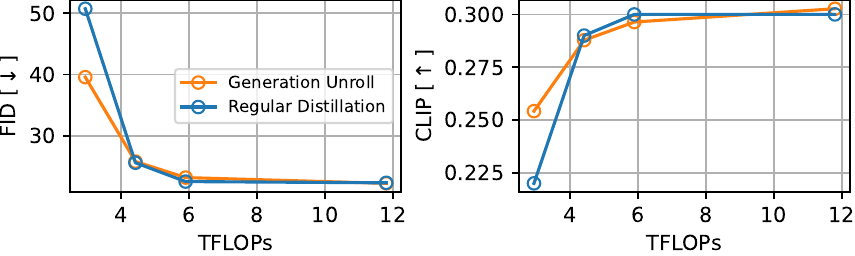}
    \caption{Training scheme ablation. We observe that our training with unrolled trajectories is generally on par with regular distillation, but performs significantly better in the low compute regime (4 steps). We use $512\times512$ MS-COCO 2017-5K, a clock of $2$ and Efficient UNet as backbone.}
    \label{fig:forward_backward}
\end{figure}

\paragraph{Adaptor Architecture.} 
We study the effect of different parametric functions for the adaptor in terms of performance and complexity.
As discussed in \cref{sub:model_step_distillation},~$\adap$ can be as simple as an identity function, where we directly reuse low-res features from the previous time step at the current step. As shown in \cref{tab:adaptor_architecture}, Identity function performs reasonably well, indicating high correlation in low-level features of the UNet across diffusion steps. In addition, we tried 1) a UNet-like convolutional architecture with two downsampling and upsampling modules, 2) a lighter variant of it with 3M parameters and less channels, 3) our proposed ResNet-like architecture (see \cref{fig:architecture}). Details for all variants are given in \cref{sec:appendix:arch_details}. From \cref{tab:adaptor_architecture}, all adaptors provide comparable performance, however, the ResNet-like adaptor obtains better quality-complexity trade-off.
\paragraph{Adaptor Clock.}
Instead of applying $\adap$ in an alternating fashion (\ie a clock of $2$), in this ablation we study the effect of non-alternating arbitrary clock $\cC(t)$. For an 8-step generation, we use 1) $\cC(t)=1$ for $t \in \{5, 6, 7, 8\}$ and 2) $\cC(t)=1$ for $t \in \{3, 4, 5, 6\}$, $\cC(t)=0$ otherwise. As shown in \cref{tab:adaptor_architecture}, both configurations underperform compared to the alternating clock, likely due to error propagation in approximation. It is worth noting that approximating earlier steps (config. 2) harms the generation significantly more than later steps (config. 1). 
\paragraph{UNet cut-off.} We ablate the splitting point where high-res and low-res representations are defined. In particular, we set the cut-off at the end of stage 1 or stage 2 of the UNet (after first and second downsampling layers, respectively). A detailed view of the architecture with splitting points can be found in the supplementary material. The lower the resolution in the UNet we set the cutoff to, the less compute we will save. As shown in \cref{tab:adaptor_architecture}, splitting at stage 2 is both more computationally expensive and worse in terms of FID. Therefore, we set the cut-off point at stage 1.
\paragraph{Training scheme and robustness.} %
As outlined in \cref{sub:adaptor_training}, the adaptor $\adap$ can be trained using 1) the regular distillation setup which employs forward noising of an image or 2) by unrolling complete sampling trajectories conditioned on a prompt. We compare the two at specific inference steps that use the same clock. \Cref{fig:forward_backward} shows that \emph{generation unroll} performs on par with regular distillation at higher inference steps (6, 8, 16), but performs significantly better at 4 steps, which is the low compute regime that our work targets.

\begin{table}[t!]
\centering
\resizebox{.9\columnwidth}{!}{
\begin{tabular}{lccccc} 
\toprule
   & \textbf{Steps} & \textbf{FID [$\downarrow$]} & \textbf{CLIP [$\uparrow$]} & \textbf{GFLOPs} \\ 
\midrule

Efficient UNet                        & 8 & 24.22 & 0.302 & 1187 \\ 
\midrule
\textbf{Adaptor Architecture} \\
\quad  Identity  (0)                   & 8 & 24.36 & 0.290 & 287 \\ 
\quad  ResNet (14M)                    & 8 & 23.21 & 0.296 & 301 \\ 
\quad  UNet (152M)                     & 8 & 23.18 & 0.296 & 324 \\ 
\quad  UNet-light (3M)                 & 8 & 23.87 & 0.294 & 289 \\ 
\midrule
\textbf{Adaptor Clock} \\
\quad  Steps $\{2, 4, 6, 8\}$                   & 8 & 23.21 & 0.296 & 301 \\
\quad  Steps $\{5, 6, 7, 8\}$                   & 8 & 28.07 & 0.286 & 301 \\
\quad  Steps $\{3, 4, 5, 6\}$                   & 8 & 33.10 & 0.271 & 301 \\
\midrule
\textbf{UNet cut-off} \\
\quad  Stage 1 (res 32x32)                   & 8 & 23.21 & 0.296 & 301 \\ 
\quad  Stage 2 (res 16x16)                   & 8 & 24.49 & 0.296 & 734\\  

\bottomrule
\end{tabular}}
\caption{Ablations of Clockwork components. We use $512\times512$ MS-COCO 2017-5K, a clock of $2$ and Efficient UNet as backbone. FLOPs are reported for 1 forward step of UNet with adaptor.
}
\label{tab:adaptor_architecture}
\vspace{-0.3 em}
\end{table}

\section{Conclusion}

We introduce a method for faster sampling with diffusion models, called \emph{\method\ Diffusion}. It combines model and step distillation, replacing lower-resolution UNet representations with more lightweight adaptors that reuse information from previous sampling steps. In this context, we show how to design an efficient adaptor architecture, and present a sampling scheme that alternates between approximated and full UNet passes. We also introduce a new training scheme that is more robust than regular step distillation at very small numbers of steps. It does not require access to an image dataset and training can be done in a day on a single GPU.
We validate our method on text-to-image generation and text-conditioned image-to-image translation \cite{tumanyan2023pnp}. It can be applied on top of commonly used models like Stable Diffusion \cite{rombach2022high}, as well as heavily optimized and distilled models, and shows consistent savings in FLOPs and runtime at comparable FID and CLIP score.

\paragraph{Limitations.} Like in step distillation, when learned, \method\ is trained for a fixed operating point and does not allow for drastic changes to scheduler or sampling steps at a later time. While we find that our unrolled trainings works better than regular distillation at low steps, we have not yet fully understood why that is the case. Finally, we have only demonstrated improvements on UNet-based diffusion models, and it is unclear how this translates to \eg ViT-based implementations.

{\small
\bibliographystyle{cvpr2023_stylekit/ieee_fullname}
\bibliography{arxiv}
}

\newpage
\appendix

\section{Clockwork details}
\label{sec:appendix:arch_details}

\begin{sidewaysfigure*}
    \centering
    \includegraphics[width=\textwidth]{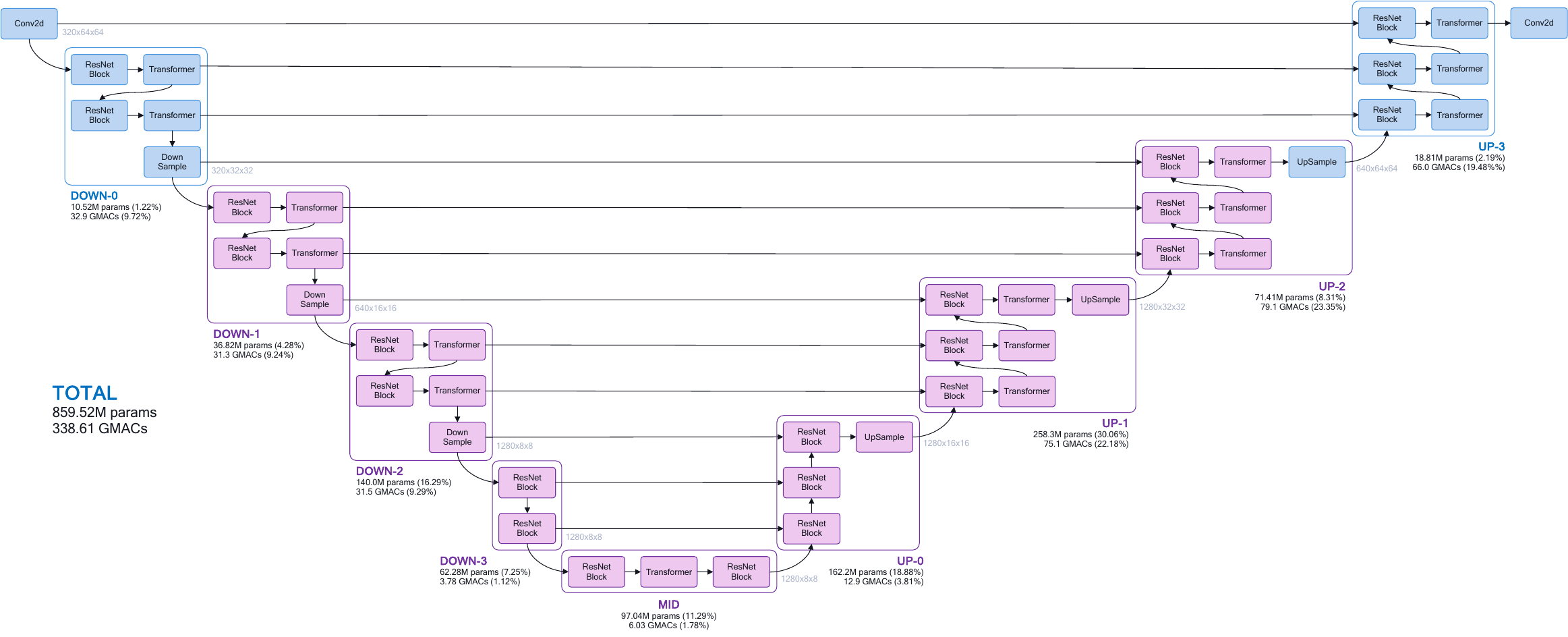}
    \caption{Detailed view of the SD UNet architecture. We replace the pink/purple parts with a lightweight adaptor, the input to which has $32\times32$ spatial resolution. For the ablations in the main body we also tried leaving DOWN-1 and UP-2 in the higher-resolution path, only replacing blocks below.}
    \label{fig:sd1.5_detailed}
\end{sidewaysfigure*}

\paragraph{UNet Architecture} In \cref{fig:sd1.5_detailed} we show a detailed schematic of the SD UNet architecture. The parts in pink are replaced by our lightweight adaptor. We also show parameter counts and GMACs per block. In ablations we varied the level at which we introduce the adaptor, as shown in Table 3 of the main body. There we compare ``Stage 1 (res 32x32)'' (our default setup) and ``Stage 2 (res 16x16)'' (a variant where DOWN-1 and UP-2 remain in the model), finding better performance for the former. Interestingly, our sampling analysis suggested that introducing the adaptor at such a high resolution, replacing most parts of the UNet, should lead to poor performance. However, this is only true if we replace multiple consecutive steps (see adaptor clock ablations in Table 3 of the main body). By alternating adaptor and full UNet passes we recover much of the performance, and can replace more parts of the UNet than would otherwise be possible.

\begin{figure}
    \centering
    \includegraphics[width=0.35\textwidth]{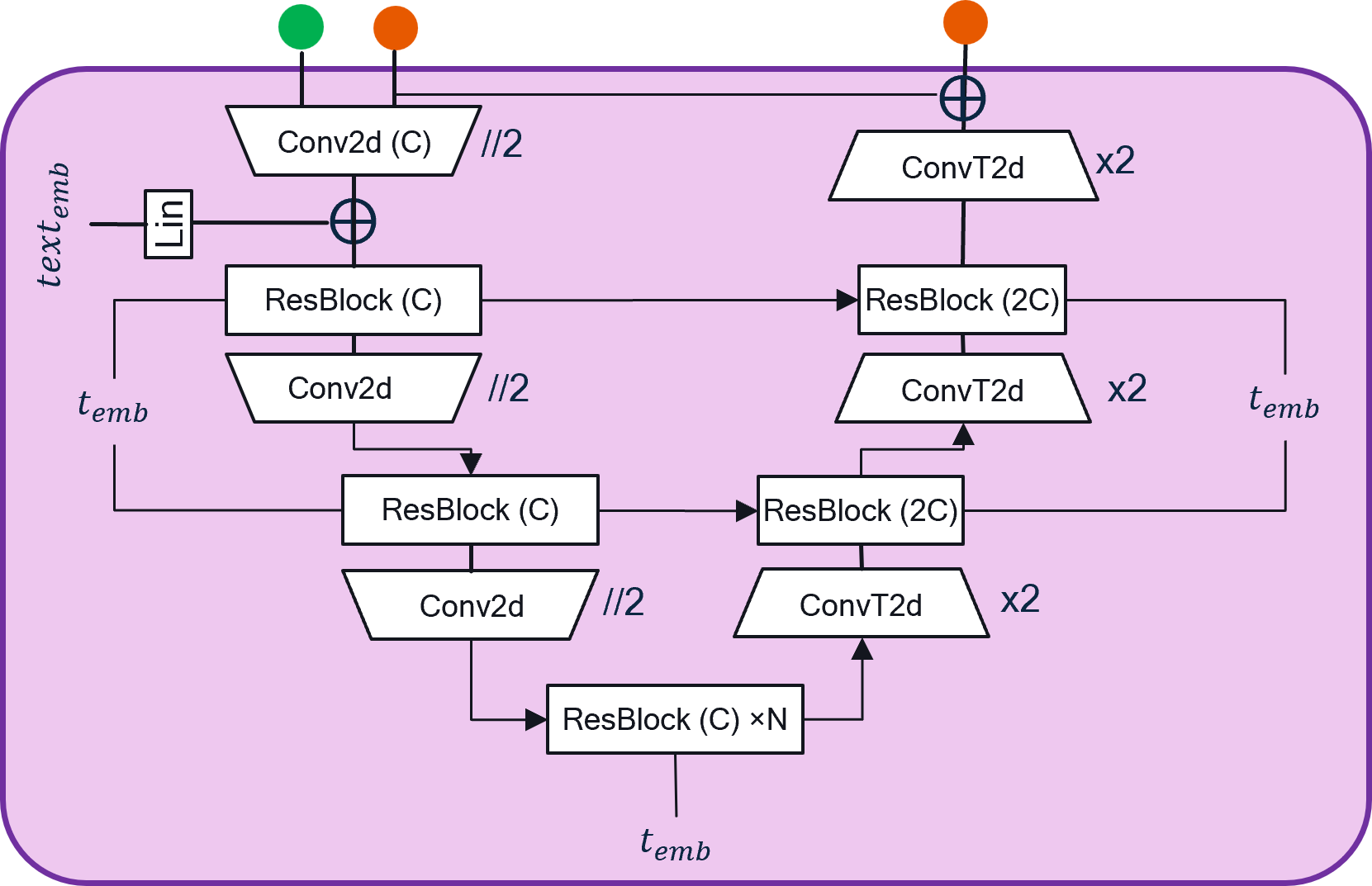}
    \caption{Architecture of a variant of the adaptor: UNet and UNet-light. For UNet we set $C=640$ and $N=4$, while for UNet-light we set $C=96$ and $N=2$.}
    \label{fig:adaptor_arch}
\end{figure}

\paragraph{Adaptor Architecture} In \cref{fig:adaptor_arch} we show a schematic of our UNet-like adaptor architecture, as discussed in ablations (Section 5.4 of the main paper). In addition to our ResNet-like architecture (Fig. 3 of the main paper) We tried 1) a UNet-like convolutional architecture with $640$ channels in each block and 4 ResNet blocks in the middle level ($N=4$), 2) a lighter variant of it with 96 channels and $2$ ResNet blocks in the middle level. While all adaptors provide comparable performance, the ResNet-like adaptor obtains better quality-complexity trade-off.

\paragraph{Training} We provide pseudocode for our unrolled training in \cref{alg:unrolled_training_alg}.

\begin{algorithm}
\caption{Adaptor training with unrolled trajectories}
\label{alg:unrolled_training_alg}
\begin{algorithmic}
    \Require Teacher model $\epsilon$
    \Require Adaptor $\adap$
    \Require Prompt set $P$
    \Require Clock schedule $\cC(t)$
    \For{$N_e$ epochs}
    \State $P_D \gets \text{RandomSubset}_D(P)$ \Comment{optional}
    \State $D \gets \text{GenerateTrajectories}(P_D, \epsilon)$
        \ForAll{Trajectory \&\ prompt $(T,text) \in D$}
            \ForAll{$(t,\repin{t},\repout{t}\repout{t+1})\in T$}
                \If{$\cC(t)=1$}
                    \State $\repoutpred{t} \gets \adap\left(\repin{t},\repout{t+1},\bm{t}_{emb},\bm{text}_{emb}\right)$
                    \State $\cL \gets \norm{\repout{t}-\repoutpred{t}}_2$
                    \State $\theta \gets \theta - \gamma\nabla\cL$
                \EndIf
            \EndFor
        \EndFor
    \EndFor
\end{algorithmic}
\end{algorithm}

\section{Ablations}
\label{sec:appendix:ablation}

\paragraph{Scheduler.} We evaluate \method\ across multiple schedulers: DPM++, DPM, PNDM, and DDIM. With the exception of DDIM, \method\ improves FID at negligible change to the CLIP score, while reducing FLOPs by $38\%$.

\begin{table}[t!]
\centering
\resizebox{.7\columnwidth}{!}{
\begin{tabular}{lcccc} 
\toprule
   & \textbf{Steps} & \textbf{FID} & \textbf{CLIP} & \textbf{TFLOPs}\\ 
\midrule
\quad DPM++                   & 8 & 24.22 & 0.302 & 9.5\\ 
\quad + Clockwork             & 8 & 23.21 & 0.296 & 5.9\\ 
\midrule
\quad DPM                     & 8 & 24.32 & 0.301 & 9.5\\
\quad + Clockwork             & 8 & 23.24 & 0.296 & 5.9\\
\midrule
\quad PNDM                    & 8 & 35.64 & 0.272 & 9.5\\
\quad + Clockwork             & 8 & 33.15 & 0.280 & 5.9\\
\midrule
\quad DDIM                    & 8 & 34.72 & 0.287 & 9.5\\
\quad + Clockwork             & 8 & 38.38 & 0.280 & 5.9\\

\bottomrule
\end{tabular}}
\caption{Clockwork works with different schedulers.
}
\label{tab:clockwork_schedulers}
\end{table}

\paragraph{Adaptor inputs.} We vary the inputs to the adaptor $\adap$. In the simplest version, we only input $\repin{t}$ and the time embedding. It leads to a poor FID and CLIP. Using only $\repout{t+1}$ provides good performance, indicating the importance of using features from previous steps. Adding $\repin{t}$ helps for a better performance, showcasing the role of the early high-res layers of the UNet.
Finally, adding the pooled prompt embedding $text_{emb}$ doesn't change FID and CLIP scores.

\paragraph{Model Distillation} In  \cref{tab:adaptor_architecture} In previous ablation, we used clock of 2. In this ablation, we explore the option to distill the low resolution layers of $\unet$ into the adaptor $\adap$, for all the steps. Here we train the adaptor in a typical model distillation setting - i.e., the adaptor $\adap$ receives as input the downsampled features at current timesteps $\repin{t}$ along with the time and text embeddings $t_{emb}$ and $text_{emb}$. It learns to predict upsampled features at current timestep $\repout{t}$. During inference, we use the adaptor during all sampling steps. Replacing the lower-resolution layers of $\unet$ with a lightweight adaptor results in worse performance.
It is crucial that the adaptor be used with a clock schedule along with input from a previous upsampled latent.

\begin{table}[t!]
\centering
\resizebox{.9\columnwidth}{!}{
\begin{tabular}{lccccc} 
\toprule
   & \textbf{Steps} & \textbf{FID [$\downarrow$]} & \textbf{CLIP [$\uparrow$]} & \textbf{GFLOPs} \\ 
\midrule

Distilled Efficient UNet                    & 8 & 25.75 & 0.297 & 150 \\ 
\midrule
\textbf{Adaptor Input} \\
\quad $r_{t}^{IN}$ + $t_{emb}$              & 8 & 40.73 & 0.262 & 150 \\
\quad $r_{t+1}^{OUT}$ + $t_{emb}$           & 8 & 24.76 & 0.295 & 150 \\ 
\quad \quad + $r_{t}^{IN}$                  & 8 & 24.45 & 0.295 & 150 \\ 
\quad \quad + $text_{emb}$                  & 8 & 24.45 & 0.295 & 150 \\ 
\textbf{Model Distillation} \\
\quad $r_{t}^{IN}$ + $t_{emb}$ + $text_{emb}$  & 8 & 117.64 & 0.06 & 150\\

\bottomrule
\end{tabular}}
\caption{\textbf{Ablation} of adaptor inputs. We use the MSCOCO-2017 dataset, Distilled Efficient UNet as backbone and a clock of $2$ (except for Model distillation where we use adaptor for all the steps) . FLOPs are reported for 1 forward step of UNet with adaptor.
}
\label{tab:adaptor_architecture}
\end{table}

\paragraph{Timings for different GPU models} In \cref{tab:timings_gpu} we report latency of different UNet backbones on different GPU models.

\begin{table}
    \centering
    \resizebox{\linewidth}{!}{
    
    \begin{tabular}{rcccc}
    \toprule
        \textbf{Latency [ms]} & RTX 3080 & RTX 2080Ti & V100 & A100 \\
    \midrule
        SD v1.5 & 454 & 589 & 453 & 235 \\
        + Clockwork & 341 & 440 & 360 & 183 \\
        & ($-24.9\%$) & ($-25.3\%$) & ($-20.5\%$) & ($-22.1\%$) \\
    \midrule
        Eff. UNet & 330 & 427 & 312 & 176 \\
        + Clockwork & 213 & 268 & 212 & 118 \\
        & ($-35.5\%$) & ($-37.2\%$) & ($-32.1\%$) & ($-33.0\%$) \\
    \midrule
        Eff. UNet (distilled) & 240 & 302 & 245 & 191 \\
        + Clockwork & 154 & 190 & 159 & 122 \\
        & ($-35.8\%$) & ($-37.1\%$) & ($-35.1\%$) & ($-36.1\%$) \\
    \bottomrule
    \end{tabular}
    
    }
    \caption{Latency improvements [ms] using Clockwork on different GPU models. All measurements are averaged over 10 runs, using DPM++ with 8 steps and batch size 1 (distilled) or 2 (for classifier-free guidance).}
    \label{tab:timings_gpu}
\end{table}

\section{Additional perturbation analyses}
\label{sec:appendix:analysis}
In Section 3 of the main body, we introduced perturbation experiments to demonstrate how lower-resolution features in diffusion UNets are more robust to perturbations, and thus amenable to distillation with more lightweight components. As a quick reminder, we mix a given representation with a random noise sample by assuming that the feature map is normal $\bm{f}\sim\cN(\mu_f,\sigma_f^2)$. We then draw a random sample $\bm{z}\sim\cN(0,\sigma_f^2)$ and update the feature map with:

\begin{equation}
    \bm{f}\leftarrow\mu_f+\sqrt{\alpha}\cdot(\bm{f}-\mu_f)+\sqrt{1-\alpha}\cdot\bm{z}
\end{equation}

\begin{figure*}
    \centering
    \includegraphics[width=\textwidth]{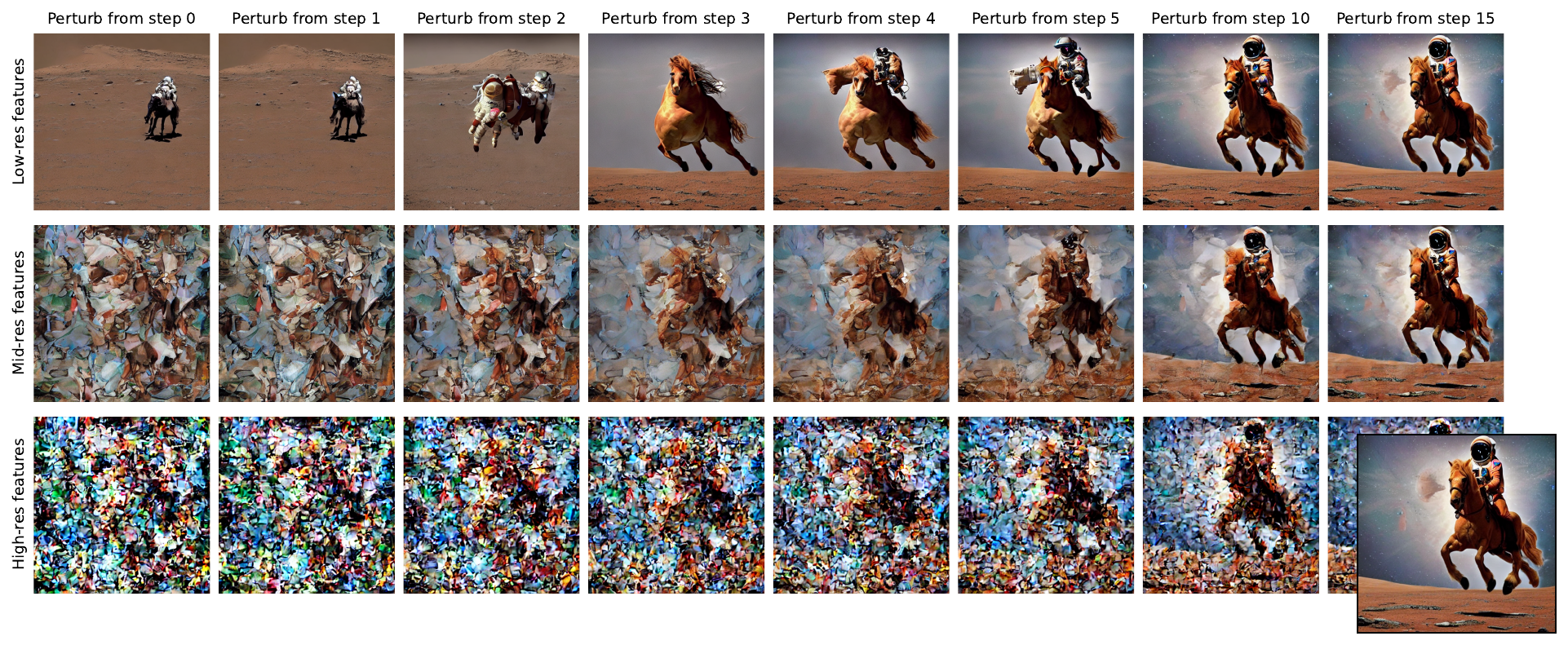}
    \caption{Reproduction of Figure 2 from the main body, using $\alpha=0.0$ (where Figure 2 uses $\alpha=0.3$). This corresponds to full perturbation of the representation, \ie the representation is completely replaced by noise in each step. Perturbation of low-resolution features still mostly results in semantic changes, whereas perturbation of higher-resolution features leads to artifacts.}
    \label{fig:perturb_quali_up_0}
\end{figure*}

For the example in the main body we set $\alpha=0.3$, so that the signal is dominated by the noise. 
Interestingly, we can also fully replace feature maps with noise, \ie use $\alpha=0.0$. The result is shown in \cref{fig:perturb_quali_up_0}. Changes are much stronger than before, but lower-resolution perturbations still result mostly in semantic changes. However, the output is of lower perceptual quality.

\begin{figure*}
    \centering
    \includegraphics[width=\textwidth]{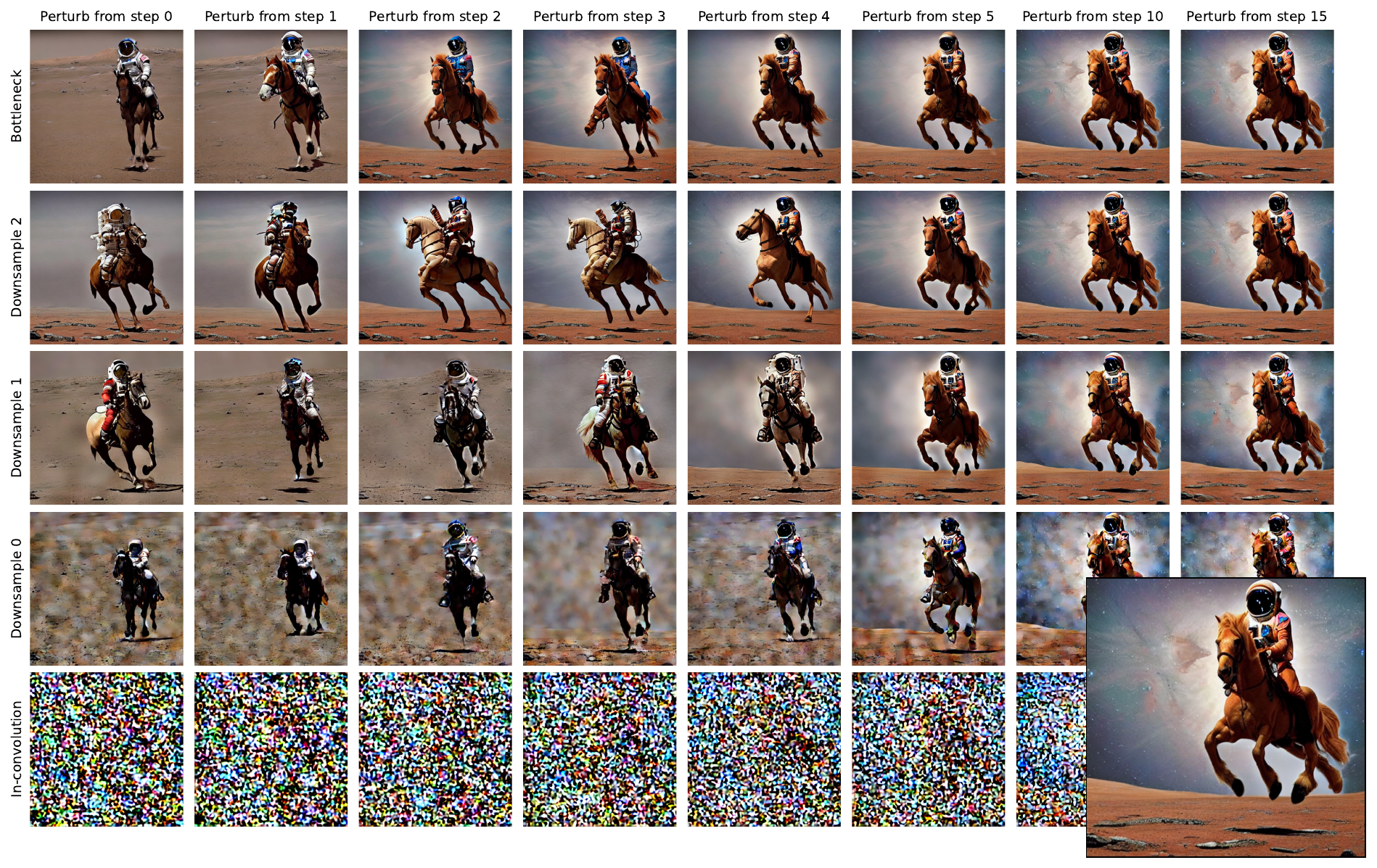}
    \caption{Reproduction of Figure 2 from the main body, perturbation different layers. Figure 2 perturbs the outputs of the 3 upsampling layers in the SD UNet, here we perturb the outputs of the 3 downsampling layers as well as the bottleneck and the first input convolution. Qualitative findings remain the same.}
    \label{fig:perturb_quali_down}
\end{figure*}

For the analysis in the main body, as well as \cref{fig:perturb_quali_up_0}, we perturb the output of the three upsampling layers in the SD UNet. We perform the same analysis for other layers in \cref{fig:perturb_quali_down}. Specifically, there we perturb the output of the bottleneck layer, the three downsampling layers, and the first convolutional layer of the network (which is also one of the skip connections). Qualitatively, findings remain the same, but perturbation of a given downsampling layer output leads to more semantic changes (as opposed to artifacts) compared to its upsampling counterpart.

\begin{figure}
    \centering
    \includegraphics[width=\linewidth]{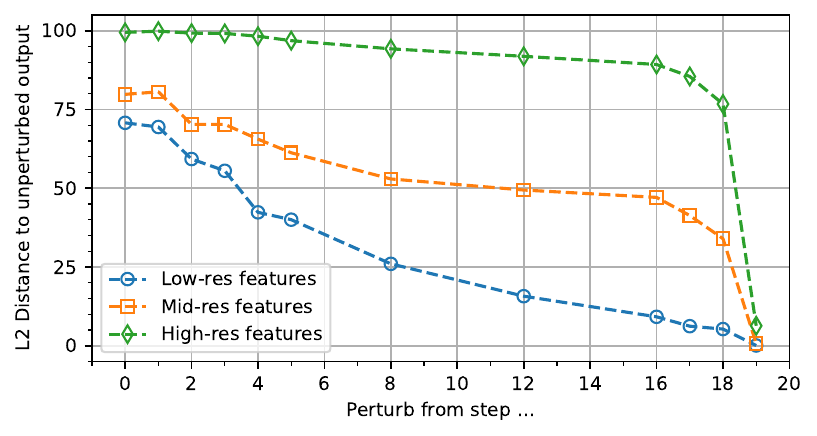}
    \caption{L2 distance to the unperturbed output, when perturbing representations with noise ($\alpha=0.7$), starting after a given number of steps. This quantifies what is shown visually in Figure 2 in the main body. Lower-resolution representations are much more robust to perturbations, and converge to the unperturbed output faster.}
    \label{fig:perturb_quant_up}
\end{figure}

\begin{figure}
    \centering
    \includegraphics[width=\linewidth]{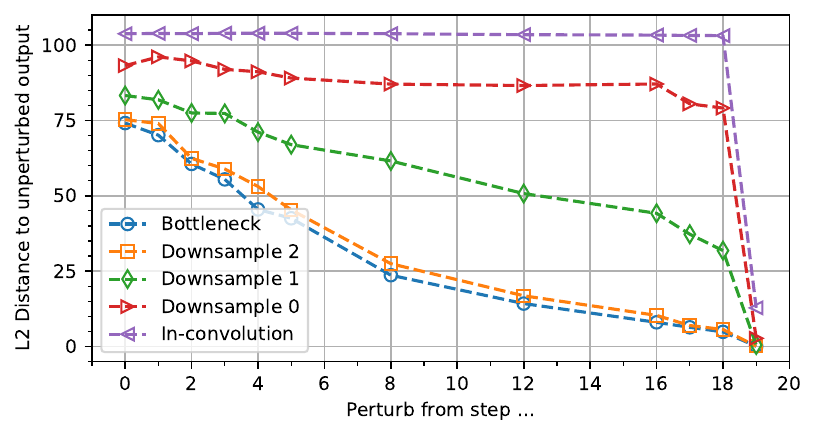}
    \caption{L2 distance to the unperturbed output, when perturbing representations with noise ($\alpha=0.7$), starting after a given number of steps. This quantifies what is shown visually in \Cref{fig:perturb_quali_down}. Lower-resolution representations are much more robust to perturbations, and converge to the unperturbed output faster.}
    \label{fig:perturb_quant_down}
\end{figure}

Finally, we quantify the L2 distance to the unperturbed output as a function of the step where we start perturbation. \cref{fig:perturb_quant_up} corresponds to the perturbations from the main body, while \cref{fig:perturb_quant_down} shows the same but corresponds to the downsampling perturbations of \cref{fig:perturb_quali_down}. Confirming what we saw visually, perturbations to low-resolution layers result in smaller changes to the final output than the same perturbations to higher-resolution features.


\section{Text-Guided Image Editing}

\subsection{Implementation Details}
\label{subsec:appendix:pnp:details}

\begin{figure*}
    \centering
    \includegraphics[trim={0, 0.3cm, 0, 0},width=0.98 \textwidth]{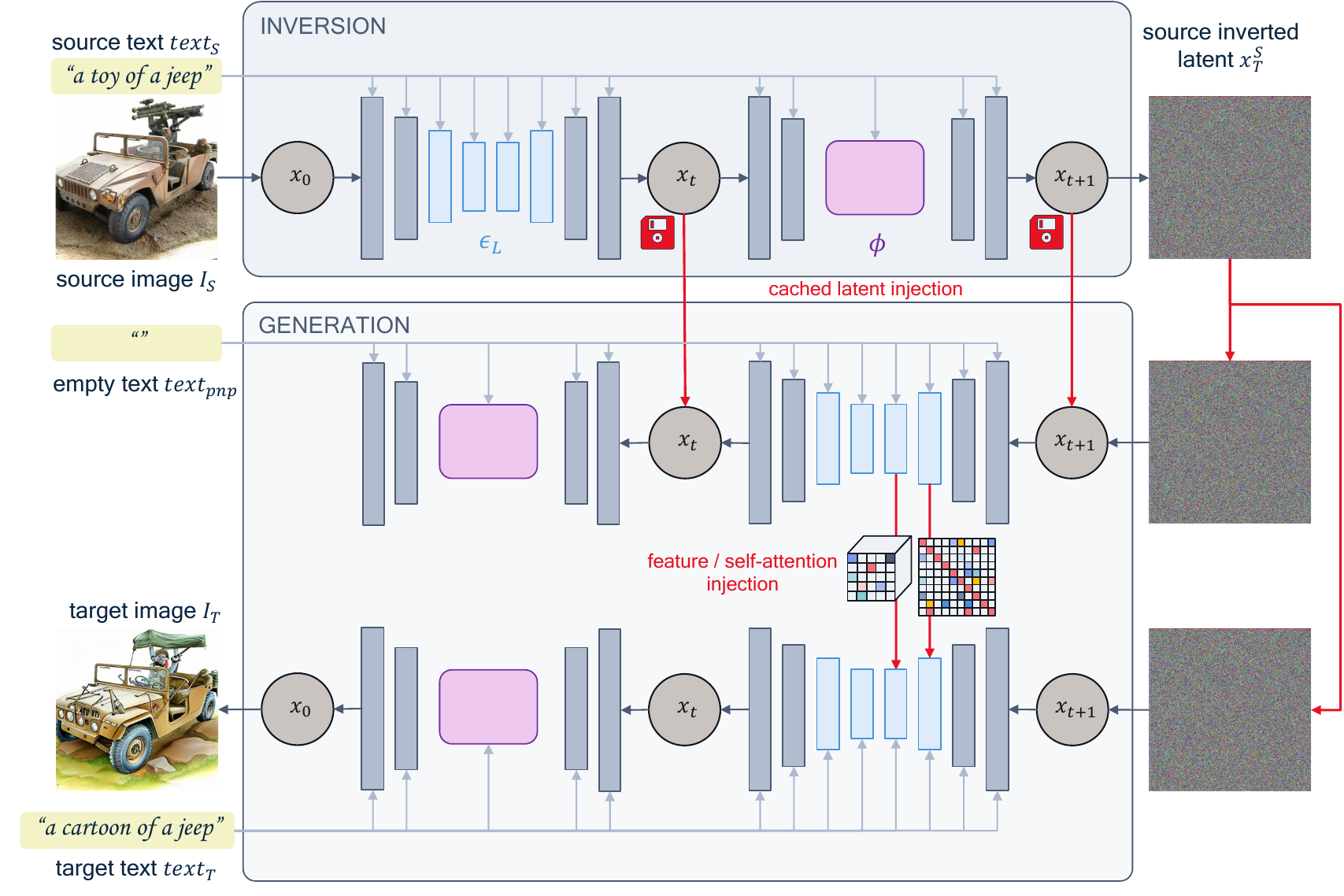}
    \caption{Overview of the actual diffusers implementation of Plug-and-Play, which contrary to what the paper describes \emph{caches latent during inversion, not intermediate features during generation}. The features to be injected are re-computed from the cached latents on-the-fly during DDIM generation sampling. The red arrows indicate injection, the floppy disk icon indicate that only the latent gets cached / saved to disk. Inversion and generation are ran separately, all operations within each are ran in-memory.}
    \label{fig:ti2i-pnp-overview}
\end{figure*}

\begin{table}[t!]
\centering
\resizebox{.85 \linewidth}{!}{
\begin{tabular}{ll} 
\toprule

\textbf{Inversion} \\
\quad  SD version                      & 1.5 \\
\quad  Sampler                         & DDIM \\
\quad  Inversion prompt                & ``a ${<}$style${>}$ of an ${<}$instance${>}$'' \\
\quad  Extract reverse                 & False \\
\midrule
\textbf{Generation} \\
\quad  SD version                      & 1.5 \\
\quad  Sampler                         & DDIM \\
\quad  Guidance scale                  & 15.0 (for both real and fake images) \\
\quad  Negative prompt                 & ``ugly, blurry, black, low res, unrealistic'' \\
\quad  $\tau_A$                        & 0.5 \\
\quad  $\tau_f$                        & 0.8 \\

\bottomrule
\end{tabular}}
\caption{Plug-and-Play hyper-parameters in inversion and generation. $\tau_A$ and $\tau_f$ are expressed as fraction of the sampling trajectory. For instance, $\tau_f=0.8$ means that for the first $80\%$ steps in the generation, convolutional features will be injected. If one uses 10 DDIM steps, this means that for the first 8 steps, convolutional features will be injected.}
\label{tab:pnp-hyper-params}
\end{table}

We base our implementation of Plug-and-Play (PnP) \cite{tumanyan2023pnp} off of the official \href{https://github.com/MichalGeyer/pnp-diffusers/tree/5d6345f4fe914993ca89765d58f50163f7b823a3}{pnp-diffusers} implementation. We summarize the different hyper-parameters used to generate the results for both the baseline and \method variant of PnP in \cref{tab:pnp-hyper-params}. Additionally, while conceptually similar we outline a couple of important differences between what the original paper describes and what the code implements. Since we use this code to compute latency and FLOP, we will go over the differences and explain how both are computed. We refer to \cref{fig:ti2i-pnp-overview} for a visual reference of the implementation of the ``pnp-diffusers''. For a better understanding, we encourage the reader to compare it to Fig. 2 from the PnP \cite{tumanyan2023pnp} paper.

\paragraph{When are features cached?} The paper describes that the source image is first inverted, and only then features are cached during DDIM sampling. They are only cached at sampling step $t$ falling within the injection schedule, which is defined by the two hyper parameters $\tau_f$ and $\tau_A$ which corresponds to the sampling steps until which feature and self-attention will be injected respectively. The code, instead of caching features during DDIM generation at time steps corresponding to injection schedule, caches during all DDIM inversion steps. This in theory could avoid running DDIM sampling using the source or no prompt. However as we will see in the next paragraph, since the features are not directly cached but the latents are, we end up spending the compute on DDIM sampling anyway.

\paragraph{What is cached?} The paper describes the caching of spatial features from decoder layers $\mathbf{f}^4_t$ along with their self-attention $\mathbf{A}^l_t$, where $4$ and $l$ indicate layer indices. The implementation trades off memory for compute by caching the latent $x_t$ instead, and recomputes the activations on the fly by stacking the cached latent along the batch axis along with an empty prompt. The code does not optimize this operation and stacks such latent irrespective of whether it will be injected, which results in a batch size of 3 throughout the whole sampling trajectory: (1) unconditional cached latent forward (2) latent conditioned on target prompt and (3) latent conditioned on negative prompt. This has implications on the latency and complexity of the solution, and we reflected it on the FLOP count, we show the formula we used in \cref{eq:flop_pnp}. 

Of note, this implementation which caches latents instead of features has a specific advantage for \method, as it enables mismatching inversion and generation steps and clock. During inversion, when the latents are cached, it does not matter whether it is obtained using a full UNet pass or an adaptor pass.
During generation, when the features should be injected, the cached latent is simply ran through the UNet to obtain the features on-the-fly. This is illustrated in \cref{fig:ti2i-pnp-overview} where features are injected at step $t+1$ during the generation although the adaptor was used at the corresponding step during inversion.

\paragraph{How do we compute FLOP for PnP?} To compute FLOP, we need to understand what data is passed through which network during inversion and generation. Summarizing previous paragraphs, we know that:
\begin{itemize}
    \item inversion is ran with a batch size of 1 with the source prompt only.
    \item generation is ran with a batch size of 3. The first element in the batch corresponds to the cached latent and the empty prompt. The other two corresponds to the typical classifier-free guidance UNet calls using the target prompt and negative prompt.
    \item both during inversion and generation, if the adaptor is used, only the higher-res part of the original UNet will be run, $\unetH$.
\end{itemize}

Let us denote $N$ and $C$ the number of steps and the clock, the indices $I$ and $G$ standing for inversion and generation respectively. We first count the number of full UNet pass in each, using integer division $N^{full}_I = N_I\ \mathbf{div}\ C_I$ (we follow similar logic for $N^{full}_G$. Additionally, we use FLOP estimate for a single forward pass with batch size of 1 in UNet,
$F_{\unet}=677.8\ \text{GFLOPs}$, and UNet with identity adaptor, $F_{\unetH+\phi}=F_{\unetH}=228.4\ \text{GFLOPs}$. The estimates are obtained using the DeepSpeed library \cite{rasley2020deepspeed}. Finally, we obtain the FLOP count $F$ as follows:

\begin{align}
    F_I &= N^{full}_I \cdot F_{\unet} + (N_I - N^{full}_I) \cdot F_{\unetH} \\
    F_G &= 3 \cdot \left(N^{full}_G \cdot F_{\unet} + (N_G - N^{full}_G) \cdot F_{\unetH}\right)\label{eq:flop_pnp} \\ 
    F &= F_I + F_G
\end{align}

\paragraph{How do we compute latency for PnP?} As described in Section 5, we only compute latency of the inversion and generation loops using \href{https://gist.github.com/sayakpaul/27aec6bca7eb7b0e0aa4112205850335}{PyTorch's benchmark utilities}. In particular, we exclude from latency computation any ``fixed'' cost like VAE decoding and text encoding. Additionally, similar to the FLOP computation, we did not perform optimization over the official PnP implementation, which leads to a batch size of 1 in the inversion loop, and a batch size of 3 in the generation loop.

\paragraph{Interplay between injection and adaptor.} The adaptor replaces the lower resolution part of the UNet $\unetL$. Based on where we split the UNet between low- and high-res, it turns out all layers which undergo injection are skipped if adaptor $\phi$ is ran instead of $\unetL$. Hence, when adaptor is ran during generation it means no features are being injected. As the number of inversion and generation steps decrease, the effect of skipping injection are more and more visible, in particular structural guidance degrades. One could look into caching and injecting adaptor features to avoid losing structural guidance. Note however that this would have no effect on complexity, and might only affect PnP + \method performance in terms of CLIP and DINO scores at lower number of steps. Since optimizing PnP's performance at very low steps was not a focus of the paper, we did not pursue this thread of work.

\paragraph{Possible optimizations.} The careful reader might understand that there are low hanging fruits in terms of both latency and FLOP optimizations for PnP. First, if memory would allow, one could cache the actual activations instead of the latent during inversion, which would allow not re-running the latent through the UNet at generation time. Second, it would be simple to modify the generation loop code \emph{not} to stack the cached latent when $t$ does not fall within the injection schedule. If implemented, a substantial amount of FLOP and latency could be saved on the generation, as the default PnP hyper parameters $\tau_f$ and $\tau_A$ lead to injection in only the first $80\%$ of the sampling trajectory. Note however that both of these optimizations are orthogonal to \method, and would benefit both the baseline and \method implementations of PnP, which is why we did not implement them.

\subsection{Additional Quantitative Results}
\label{sec:appendix:pnp:quantitative}

\begin{figure*}
    \centering
    \includegraphics[trim={0, 0.3cm, 0, 0},width=0.98 \textwidth]{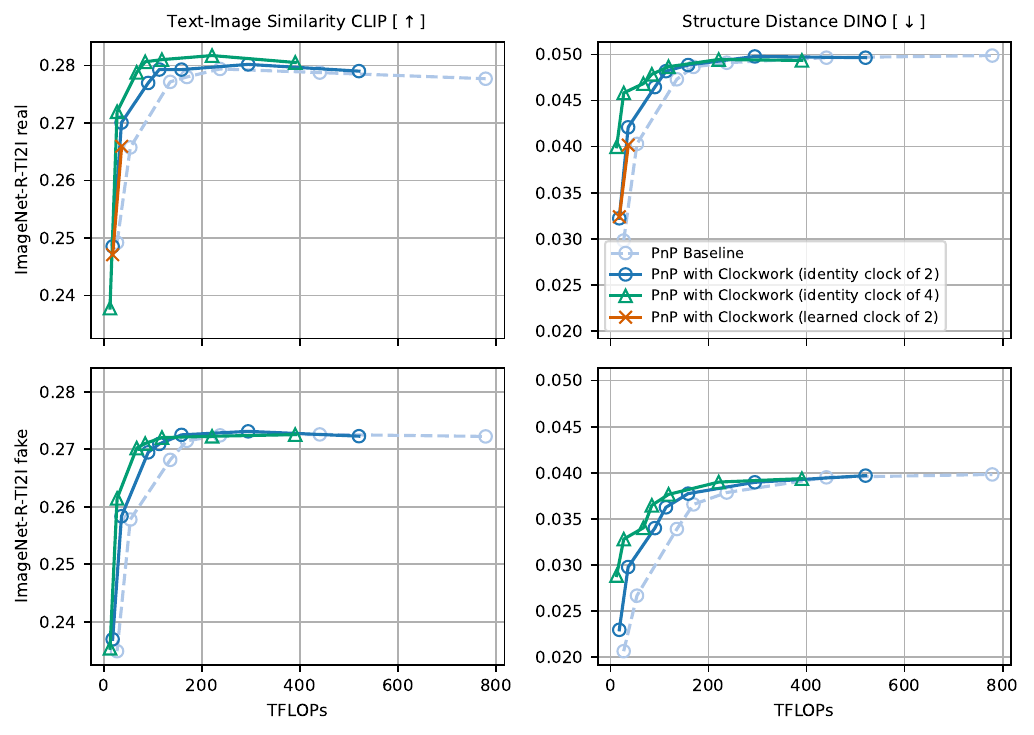}
    \caption{Additional quantitative results on ImageNet-R-TI2I real (top) and fake (bottom) for varying number of DDIM inversion steps: $[10, 20, 25, 50, 100, 200, 500, 1000]$. We use $50$ generation steps except for inversion steps below 50 where we use the same number for inversion and generation.}
    \label{fig:ti2i-pnp-quantitative-appendix}
\end{figure*}

We provide additional quantitative results for PnP and its \method variants. In particular, we provide CLIP and DINO scores at different clocks and with a learned ResNet adaptor. In addition to the ImageNet-R-TI2I \emph{real} dataset results, we report scores on ImageNet-R-TI2I \emph{fake} \cite{tumanyan2023pnp}.

In the \cref{fig:ti2i-pnp-quantitative-appendix}, we can see how larger clock size of $4$ enables bigger FLOP savings compared to $2$, yet degrade performance at very low number of steps, where both CLIP and DINO scores underperform at 10 inversion and generation steps. It is interesting to see that the learned ResNet adaptor does not outperform nor match the baseline, which is line with our ablation study which shows that \method works best for all schedulers but DDIM at very low number of steps, see \cref{tab:clockwork_schedulers}.

We can see that results transfer well across datasets, where absolute numbers change when going from ImageNet-R-TI2I real (top row) to fake (bottom row) but the relative difference between methods stay the same.

\subsection{Additional Qualitative Results}

\begin{figure*}
    \centering
    \includegraphics[trim={0, 0.3cm, 0, 0},width=0.98 \textwidth]{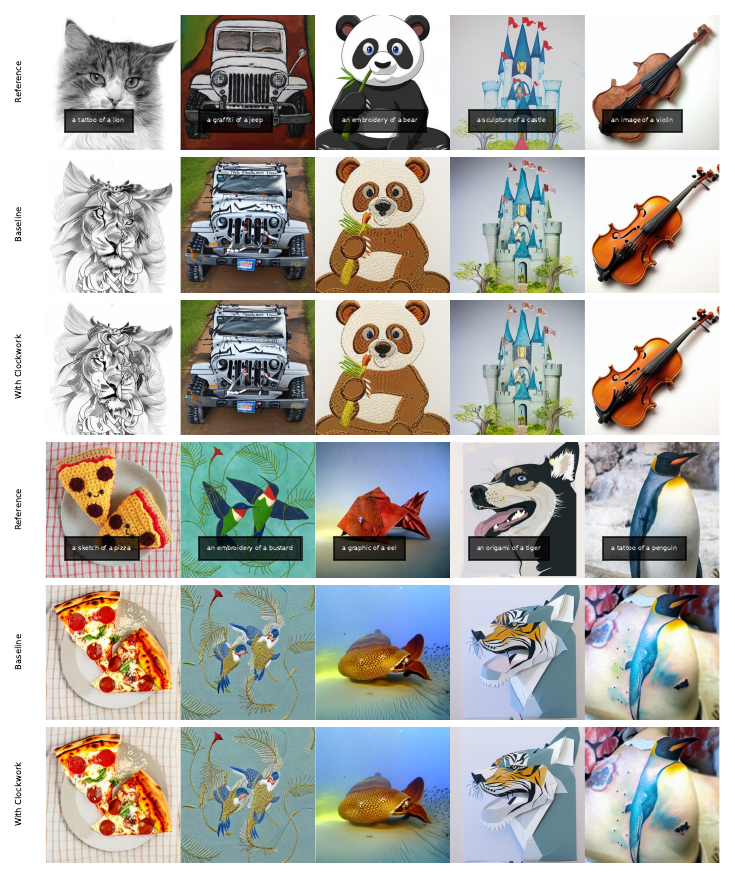}
    \caption{Examples from ImageNet-R-TI2I \emph{real} from Plug-and-Play~\cite{tumanyan2023pnp} and its \method variant. We use 50 DDIM inversion and generation steps, and a clock of $2$. Images synthesized with \method are generated $34\%$ faster than the baseline, while being perceptually close if at all distinguishable from baseline.}
    \label{fig:ti2i-pnp-qualitative-real}
\end{figure*}

\begin{figure*}
    \centering
    \includegraphics[trim={0, 0.3cm, 0, 0},width=0.98 \textwidth]{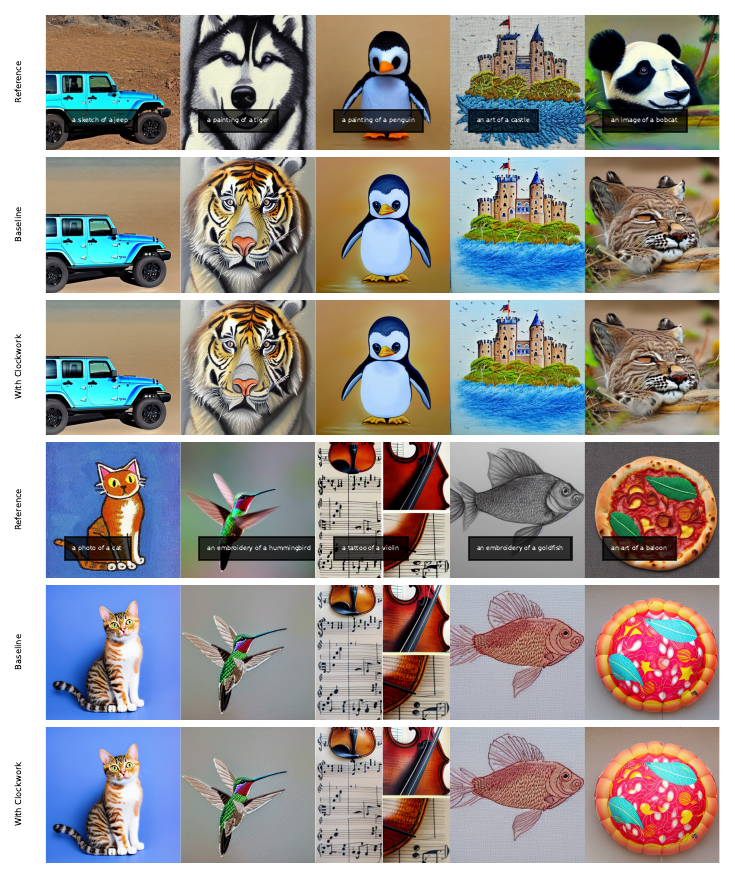}
    \caption{Examples from ImageNet-R-TI2I \emph{fake} from Plug-and-Play~\cite{tumanyan2023pnp} and its \method variant. We use 50 DDIM inversion and generation steps, and a clock of $2$. Images synthesized with \method are generated $34\%$ faster than the baseline, while being perceptually close if at all distinguishable from baseline.}    \label{fig:ti2i-pnp-qualitative-fake}
\end{figure*}

We provide additional qualitative examples for PnP for ImageNet-R-TI2I real in \cref{fig:ti2i-pnp-qualitative-real} and \cref{fig:ti2i-pnp-qualitative-fake}. We show examples at 50 DDIM inversion and generation steps.

\section{Additional examples}
\label{sec:appendix:t2i_examples}

\begin{figure*}
    \centering
    \includegraphics[width=\textwidth]{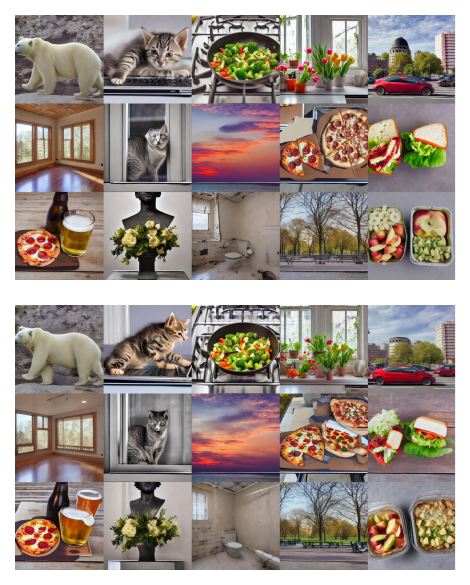}
    \caption{Additional example generations for SD UNet without (top) and with (bottom) \method. We include the examples shown in the main body so that the layout of this figure matches that of \cref{fig:examples_efficientunet} and \cref{fig:examples_efficientunet_distill}.}
    \label{fig:examples_sdunet}
\end{figure*}

\begin{figure*}
    \centering
    \includegraphics[width=\textwidth]{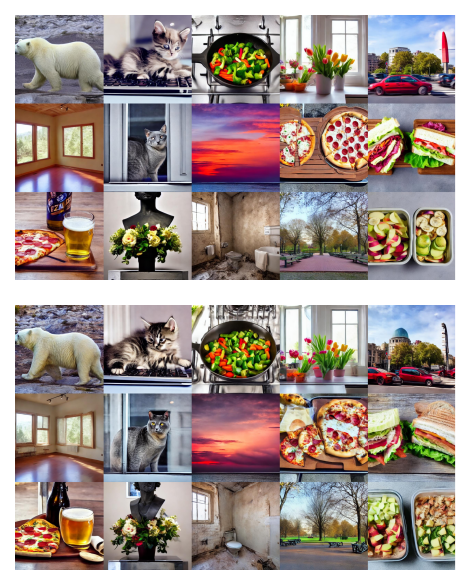}
    \caption{Example generations for Efficient UNet without (top) and with (bottom) \method.}
    \label{fig:examples_efficientunet}
\end{figure*}

\begin{figure*}
    \centering
    \includegraphics[width=\textwidth]{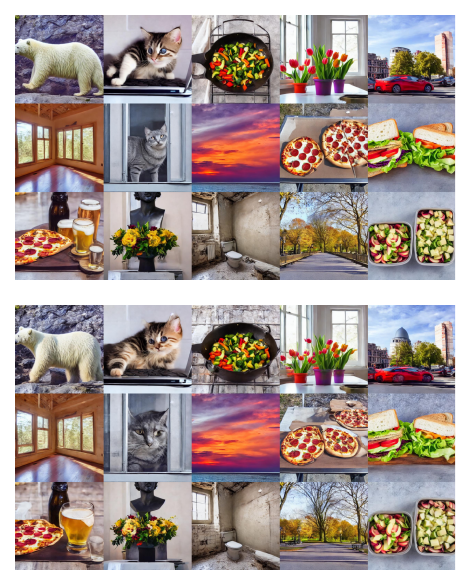}
    \caption{Example generations for Distilled Efficient UNet without (top) and with (bottom) \method.}
    \label{fig:examples_efficientunet_distill}
\end{figure*}

We provide additional example generations in this section. Examples for SD UNet are given in \cref{fig:examples_sdunet}, examples for Efficient UNet in \cref{fig:examples_efficientunet}, and those for the distilled Efficient UNet in \cref{fig:examples_efficientunet_distill}. In each case the top panel shows the reference without \method\ and the bottom panel shows generations with \method. \cref{fig:examples_sdunet} includes the same examples already shown in the main body so that the layout is the same as for the other models for easier comparison.

The prompts that were used for the generations are the following (left to right, top to bottom), all taken from the MS-COCO 2017 validation set:

\begin{itemize}
\item ``a large white bear standing near a rock.''
\item ``a kitten laying over a keyboard on a laptop.''
\item ``the vegetables are cooking in the skillet on the stove.''
\item ``a bright kitchen with tulips on the table and plants by the window ''
\item ``cars waiting at a red traffic light with a dome shaped building in the distance.''
\item ``a big, open room with large windows and wooden floors.''
\item ``a grey cat standing in a window with grey lining.''
\item ``red clouds as sun sets over the ocean''
\item ``a picnic table with pizza on two trays ''
\item ``a couple of sandwich slices with lettuce sitting next to condiments.''
\item ``a piece of pizza sits next to beer in a bottle and glass. ''
\item ``the bust of a man's head is next to a vase of flowers.''
\item ``a view of a bathroom that needs to be fixed up.''
\item ``a picture of some type of park with benches and no people around.''
\item ``two containers containing quiche, a salad, apples and a banana on the side.''
\end{itemize}

\end{document}